\documentclass[10pt,twocolumn,letterpaper]{article}

\usepackage{cvpr}              % To produce the CAMERA-READY version
\usepackage{multirow}
\usepackage{amssymb}
\usepackage{amsmath}
\usepackage{placeins}
\usepackage{booktabs}
\usepackage{subcaption}
\usepackage[accsupp]{axessibility}

\definecolor{cvprblue}{rgb}{0.21,0.49,0.74}
\usepackage[pagebackref,breaklinks,colorlinks,allcolors=cvprblue]{hyperref}

\def\paperID{1586} % *** Enter the Paper ID here
\def\confName{CVPR}
\def\confYear{2026}

\title{Cross-Modal Attention Calibration for LVLM Hallucination Mitigation}

\author{Jiaming Li$^{1}$\thanks{Equally-contributed authors.} \,\thanks{Work done during an internship at
Meituan.}\quad
Jiacheng Zhang${^{2*}}$\quad
Zequn Jie$^{3}$\quad
Lin Ma$^{3}$\quad
Ming Li$^{4}$\quad
Xiaonan Luo$^{5}$\quad
Guanbin Li $^{1,6,7}$\thanks{Corresponding author is Guanbin Li.}\\
$^{1}$ Sun Yat-sen University \quad
$^{2}$The University of Hong Kong \quad $^{3}$Meituan \\ 
$^{4}$ Inspur Database Technology \quad 
$^{5}$ Guilin University of Electronic Technology \\
$^{6}$ Shenzhen Loop Area Institute \quad $^{7}$Guangdong Key Laboratory of Big Data Analysis and Processing \\
 {\tt\small lijm48@mail2.sysu.edu.cn, liguanbin@mail.sysu.edu.cn }\\
}

\begin{document}
\maketitle
\begin{abstract}

Large vision-language models (LVLMs) have shown remarkable capabilities in visual-language understanding. Despite their success, LVLMs still suffer from generating hallucinations in complex generation tasks, leading to inconsistencies between visual inputs and generated content. To address this issue, some approaches have introduced inference-time interventions, such as contrastive decoding, to reduce overreliance on language priors.   However, these approaches overlook hallucinations stemming from position bias and spurious inter-modality correlations. In this paper, we propose a Cross-Modal Attention Calibration (CMAC) method to mitigate hallucinations in LVLMs in a training-free manner. In this method, we design an Inter-Modality Decoding (IMD) module to alleviate hallucination by a novel contrastive decoding mechanism. IMD masks the value vectors associated with significant cross-modal attention weights as distortion, which addresses both uni-modality overreliance and misleading inter-modality correlations. Additionally, a Cross-Modal Position Calibration (CMPC) module shrinks the position gap of image tokens, alleviating the position bias in cross-modal attention. Experimental results on diverse hallucination benchmarks validate the superiority of our method over existing state-of-the-art techniques in reducing hallucinations for LVLM. Our code will be available at \url{https://github.com/lijm48/IMCCD}.
 
\end{abstract}
\section{Introduction}
\label{sec:intro}

 With advances in computational power and data availability, large language models~\cite{touvron2023llama,touvron2023llama2,chiang2023vicuna,bai2023qwen,achiam2023gpt} (LLMs) have achieved significant progress in language understanding\cite{liguided,li2024learning}, generation\cite{huang2025dreamlayer}, and reasoning\cite{wang2025free}. 
Large vision-language models~\cite{chen2023shikra,bai2023qwenvl,chen2024internvl,chen2024far,liu2024visual,liu2024improved} (LVLMs) further extend large language models to vision-language tasks, demonstrating impressive performance across a range of applications, including image captioning and visual question answering (VQA).
 Despite these advancements, LVLMs suffer from the issue of hallucinations while generating the response, in which LVLMs generate textual content that is semantically coherent but inconsistent with ground-truth objects in the given image,  hindering their reliable application.
\begin{figure}[t]
\centering
% \vspace{-0.25cm}
% \includegraphics[width=\textwidth, trim=0 0 0 5, clip]{images/pipeline_final.pdf}
\includegraphics[width=0.99\linewidth]{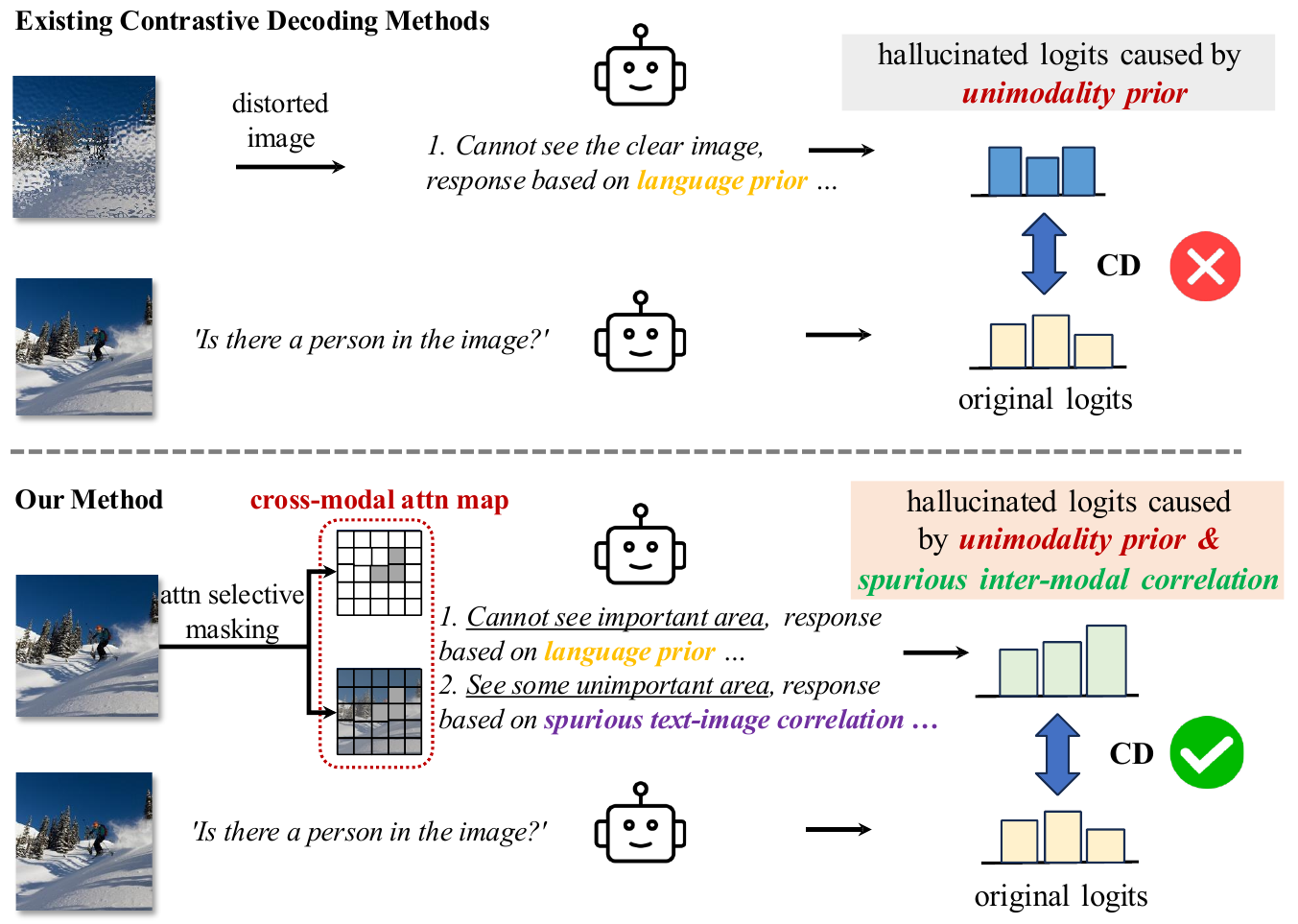}
% \vspace{-3mm}
% \vspace{-0.25cm}
\caption{ 
% The figure shows the attention weights from text tokens are biased toward the latter part of images, leading to the position bias.  Besides, the hallucination rates(estimated by recall) of objects in image tokens are highly related to the position, showing the position bias from current LVLMs.   (b) In contrast to previous contrastive decoding methods, our decoding method alleviates both language prior and spurious inter-modality correlation by selective distortion on attention.
 Illustration of the comparison between the commonly used contrastive decoding strategy method and our proposed \textit{Inter-Modality Contrastive Decoding}.
}
\label{fig:teaser}
\end{figure}

\begin{figure}[t]
\centering
\includegraphics[width=\linewidth]{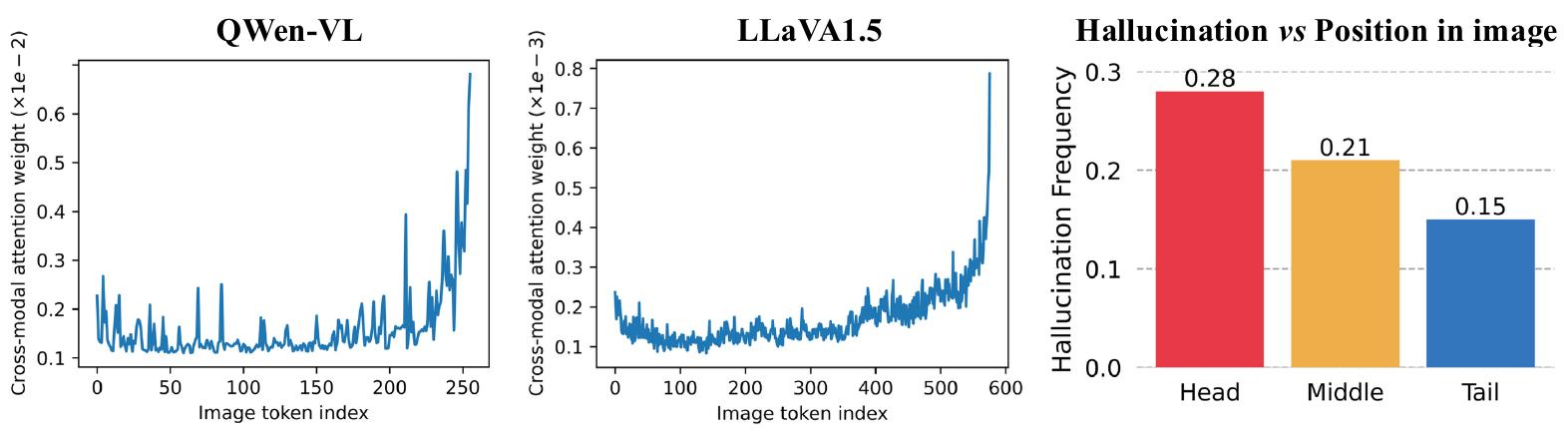}
\caption{ 
 Illustration of the cross-modal attention \textit{Position Bias} within some common LVLMs and the resulting spatially biased hallucination patterns.
}%\vspace{-3mm}
\label{fig:teaser_pos}
\end{figure}

Among various mitigation strategies, inference-time methods, particularly Contrastive Decoding (CD)\cite{leng2024mitigating, favero2024multi, woo2024don}, have gained prominence due to their training-free nature and effectiveness. The core principle of CD is to penalize tokens that are likely under a "hallucinatory" distribution, typically induced by perturbing the input (e.g., distorting visual features or masking text). While promising, we argue that existing CD-based approaches suffer from two fundamental, previously overlooked limitations.
First, they fail to disentangle genuine visual grounding from spurious cross-modal correlations. LVLMs often learn superficial associations from web-scale data, such as the co-occurrence of "food" and "dining table"~\cite{leng2024curse}. Consequently, a query about a dining table can trigger a hallucinated response even if one is absent, merely because food is present in the image. Current CD techniques, which distort the entire visual input, cannot isolate and penalize these spurious associations. The resulting "hallucinatory" distribution is too coarse, failing to provide the fine-grained contrast needed to suppress such errors.
Second, we identify a systematic positional bias in the cross-modal attention mechanism of contemporary LVLMs. As illustrated in Figure~\ref{fig:teaser_pos}, we observe that hallucinations are more likely to involve objects corresponding to the initial portion of the image token sequence. With further investigation, we reveal that the issue arises from the Rotary Position Embedding (RoPE) in the LVLM decoder. Typically, images are embedded into tokens, interleaved with text tokens, and processed with RoPE for autoregressive generation. This architecture inherently incentivizes the model to disproportionately attend to the latter parts of the visual sequence during autoregressive generation, which is demonstrated by the relatively higher attention weights of the later part of image tokens.

To overcome these challenges, we introduce the \textbf{C}ross-\textbf{M}odal \textbf{A}ttention \textbf{C}alibration (CMAC) framework for more effective hallucination mitigation in a training-free manner. 

To address spurious correlations, an \textit{ Inter-Modality Decoding (IMD)} module is proposed to craft a more effective "hallucinatory" distribution for contrastive decoding.
As illustrated in Figure~\ref{fig:teaser}, instead of globally distorting the image, \textit{IMD operates at the \textbf{attention level}}. It selectively masks the value vectors corresponding to high cross-modal attention scores. This surgical intervention suppresses strong, legitimate inter-modality correlations in cross-modal attention while leaving weak, spurious correlations and the unimodal reliance intact in the contrastive distribution.  This allows the resulting contrastive distribution to effectively capture failures from both spurious correlations and unimodal language priors.

To counteract positional bias, we propose the \textit{Cross-Modal Position Calibration (CMPC)} module, which mitigates the tendency to overlook crucial visual tokens caused by position bias. This is achieved by linearly reducing the positional index gap of image tokens in cross-modal attention, preserving their global position within sentences and alleviating positional bias.
Extensive experiments on various LVLM hallucination benchmarks demonstrate the superiority of our CMAC framework, significantly outperforming existing methods. Our main contributions are summarized as follows:

\begin{itemize}
    \item  We propose a novel  Inter-Modality Decoding method for LVLM hallucination mitigation, which eliminates both uni-modality overreliance and spurious inter-modality correlations by contrastive decoding.
    
    \item  We design a Cross-modal Position Calibration, which alleviates the position bias by scaling the position indexes of image tokens in the cross-modal attention process.
    % improving the attention to the content of visual input.
    
    \item  Comprehensive experiments on various benchmarks demonstrate the effectiveness and generalization of our proposed method in reducing hallucinations for LVLMs.
    % \item  We propose a novel  Inter-Modality Decoding method for LVLM hallucination mitigation, which eliminates both uni-modality overreliance and spurious inter-modality correlations by contrastive decoding.
    
    % \item  We design a Cross-modal Position Calibration, which alleviates the position bias by scaling the position indexes of image tokens in the cross-modal attention process.
    
    % \item  Comprehensive experiments on various benchmarks demonstrate the effectiveness and generalization of our proposed method in reducing hallucinations for LVLMs.
\end{itemize}

\section{Related Work}
\noindent \textbf{Large Vision-Language models.}
Large vision-language models (LVLMs) equip large language models (LLMs) with the capability to perceive and understand both textual input and visual input data. 
The most common practice to achieve LVLMs is to integrate pre-trained LLMs with additional visual encoders and cross-modal interfaces for cross-modal fusion. For example, the LVLMs~\cite{chen2023shikra,bai2023qwenvl,chen2024internvl,chen2024far} represented by InternVL~\cite{chen2024internvl} introduce linear projections to map the image features from the vision encoder to the token space of LLMs. While LLaVA series~\cite{liu2024visual,liu2024improved} develop a vision-language model by connecting a vision encoder and an LLM with a projection layer and fine-tuning their model based on their generated instructional vision-language data. Instead, another line of work~\cite{zhu2023minigpt, ye2023mplug, ye2024mplug2,li2023blip} such as BLIP-2~\cite{li2023blip} adopts the design of query transformer as the interface between vision encoders and LLMs. \citet{Instructblip} further proposes InstructBLIP, which enhances visual comprehension through vision-language instruction tuning and introduces instruction-aware visual feature extraction to query transformers to enable context-relevant processing of visual content based on the given instructions. Despite these advancements in LVLMs, these models continue to struggle with severe hallucination issues, where generated content misaligns with the visual inputs. Our work aims to mitigate the hallucination of current LVLMs and facilitates the application of LVLMs in various domains.

~\\
\noindent \textbf{Hallucination in LVLMs.}
The hallucination problem was first discovered in the field of LLMs. It refers to the misalignment between the generated content from LLMs and real-world facts (namely the factuality hallucination) or user instructions (namely the faithfulness hallucination). Built upon LLMs,  LVLMs also suffer from the hallucination that manifests as a misalignment between generated text and the visual input. Various approaches have been proposed to tackle this issue, including the works from the perspective of constructing additional robust training instruction~\cite{yue2024less,liu2023mitigating,yu2024hallucidoctor,jiang2024hallucination}, reinforcement learning with human/AI feedback~\cite{gunjal2024detecting,yu2024rlhf,sun2023aligning,yu2024rlaif,li2023silkie,kim2025exploiting,yang2025mitigating},  or model structure enhancement~\cite{zhai2023halle}, etc. Despite achieving impressive results, these methods always require extensive data collection or additional fine-tuning of LVLMs, making them both computationally and labor-intensive. Another line of method~\cite{wu2024logical,huang2024opera,yin2023woodpecker,zhou2023analyzing,woo2024ritual,liu2024paying,chen2025ict,an2025mitigating} focuses on the training-free method by improving the inference process of LVLMs to suppress hallucination. For example, some methods~\cite{leng2024mitigating,favero2024multi,woo2024don,huo2024self,wang2024mitigating,chen2024halc,suo2025octopus} mitigate the hallucination via reducing the over-reliance of LVLMs on the language prior by performing conservative decoding on the original inputs and the inputs with disturbed contents. To the end, other methods like VCD~\cite{leng2024mitigating} propose to distort the visual content by noise-adding~\cite{leng2024mitigating} or token-wise pruning~\cite{favero2024multi,woo2024don}, while ICD~\cite{wang2024mitigating}  implicitly distorts the instruction by introducing the negative role prefix for LVLMs.  
% Apart from contrastive decoding, PAI~\cite{liu2024paying} proposes to directly enhance the attention weights of the visual part. 
However, the distortion of textual/visual content intervenes in cross-modal attention, neglecting hallucinations caused by spurious inter-modality correlations. Additionally, the position bias is also under-explored in the training-free manner.
% \chaowei{
%  变量命名原则：

%  - 集合： 大写黑板加粗正体，e.g., $\mathbb X$

%  - 2维以上张量: 大写加粗正体，e.g., $\mathbf X$

%  - 1维向量： 小写加粗正体，e.g., $\mathbf x$

%  - 标量：不加粗斜体，e.g., $x$

%  - 函数：不加粗正体，e.g., $\textrm{ABC}(x)$, $\textrm{F}(x)$

%  - 上标为数字时，添加括号避免和指数运算冲突, e.g., $x^{(1)}$

%  - 关键变量首次出现或未按照以上规则命名时，标明其维度, e.g., $\mathbf X \in \mathbb R^{a\times b\times c}$
%  }

 % $\mathbf X = [\mathbf x_i]_{i=0}^{n}$
\section{Methodology}

\subsection{Preliminaries}
Modern LVLMs typically comprise a visual encoder, a cross-modal interface, and a language decoder. The visual encoder is usually adapted from a pre-trained vision model, while the language decoder is derived from a pre-trained LLM. During inference, LVLMs process both visual and textual inputs to generate the next token in the response sequence iteratively. Specifically, the visual encoder first encodes the input visual content into visual features. The cross-modal interface then maps these visual features to the input space of the language decoder to generate the image tokens $\mathbf{X} = \{x_i \}_{i=0}^{n}$. Here $x_i$ is the image token corresponding to the $i$-th patch of the image and $n$ is the number of patches. 
% $[\cdot]$ denotes the concentration operation. 
Besides, the text input is mapped to the text tokens $\mathbf T = \{t_i\}_{i=0}^{m}$, where $t_i$ is the $i$-th text token and $m$ is the number of text tokens.
The image tokens and text tokens are then concatenated to serve as the input tokens $\left[\mathbf T_{0:m_{b}}, \mathbf X, \mathbf T_{m_{b}+1:m}\right]$ for the language decoder. $\mathbf T_{0:m_{b}} = \left\{t_i\right\}_{i=0}^{m_{b}}$  is the first $m_{b}$ tokens (system prompt) in $\mathbf T$ and $\mathbf T_{m_{b}+1:m} = \left\{t_i\right\}_{i=m_{b}+1}^{m}$ is the remaining part of $\mathbf T$ (query prompt).

\noindent \textbf{Language Decoder Forwarding}. The multi-head self-attention layer is commonly used in the language decoder. During the forward pass of the language decoder, the input tokens are initially transformed into input embeddings by the embedding layer, which then serve as the hidden states for the first self-attention layer. For each sample, every head within a self-attention layer maps the hidden states to queries  $\mathbf Q \in \mathbb R^{(n+m) \times d}$, keys $\mathbf K \in \mathbb R^{(n+m) \times d}$, and values $\mathbf V \in \mathbb R^{(n+m) \times d}$ by linear transformation. Here $(n+m)$ is the sequent length of input tokens and $d$ denotes the hidden dimensions. 
Then the position embeddings are applied to the queries and keys to incorporate the position information of each token. Recent language decoders primarily rely on the rotary position embeddings (ROPE) to encode the position information, which estimate rotary matrices $\mathbf R$ for query and key states by, 
% \begin{equation}
%  \\
% \label{eq:rope}
% \end{equation}
\begin{align}
\mathbf{R}_k^i = 
\begin{pmatrix}
% \setstacktabbedgap{2pt}
\cos{(\mathbf P[i]\theta_k)}& -\sin{(\mathbf P[i]\theta_k)}\\
\sin{(\mathbf P[i]\theta_k})&\cos{(\mathbf P[i]\theta_k)} 
\end{pmatrix},  \quad  & 
\theta_k = \frac{1}{b^{2k/d}} \nonumber \\
\mathbf q_{\textrm p}^i =\mathbf q^i \mathbf{R}^i, \quad  \mathbf k_{\textrm  p}^i =\mathbf k^i \mathbf{R}^i, & 
\label{eq:rope}
\end{align}
Here $b$ is a hyper-parameter. $\mathbf{R}^i_k$ is the rotary matrix for the $2k$-th and $(2k+1)$-th dimension of query $\mathbf q^i$ and key states $\mathbf k^i$ of $i$-th token with position index $\mathbf P[i]$. Normally, the position indexes of all the tokens are set to $\mathbf P = [\{i\}_{i=1}^{m+n}]$. Given the queries $\mathbf Q_{\textrm  p}$ and keys $\mathbf K_{\textrm  p}$ with position embeddings, the attention matrix $\mathbf A \in \mathbb R^{(n+m)\times (n+m)}$ is estimated by,
\begin{equation}
\mathbf{A}^{l} = \frac{\mathbf{Q_{\textrm  p}K_{\textrm  p}}^T}{\sqrt{d}},\quad \mathbf A =  \textrm{softmax}(\mathbf{A}^l), \\
\label{eq:attention_weights}
\end{equation}
where $\mathbf{A}^{l}$ is the attention logits before softmax.
The attention matrix $\mathbf A$ estimates the relevance of each token, which is used to reweight the values $\mathbf V$  from each token to obtain the attention output $\mathbf O \in \mathbb R^{(n+m) \times d}$,
\begin{equation}
\mathbf O =  \mathbf{AV}, \\
\label{eq:attention_output}
\end{equation}
The attention output $\mathbf O$ is subsequently fed into the fully connected feed-forward network (FFN), whose output then serves as the input hidden states for the subsequent multi-head self-attention layer. 

\noindent \textbf{Next Token Prediction}.
The language decoder parameterized by $\theta$ in LVLMs generates the $t$-th output token based on the text tokens $\mathbf T$, image tokens $\mathbf X$, and the previously generated tokens $\mathbf Y_{<t}$ in an auto-regressive manner. This process can be formulated as follows,  
\begin{align}
\mathbf Y_t &\sim p_{\theta}(\mathbf Y_t|\mathbf T, \textrm{Attn}(\mathbf O | \mathbf T, \mathbf X, \mathbf Y_{<t})), \nonumber \\
            &\propto \textrm{logit}_{\theta}(\mathbf Y_t|\textrm{Attn}(\mathbf O | \mathbf T, \mathbf X, \mathbf Y_{<t})),
\end{align}
where  $p_{\theta}(\cdot)$  and $\textrm{logit}_{\theta}(\cdot)$ denote the output probability scores and logits of the language decoder, respectively. $\textrm{Attn}$ is the self-attention process by Eq~(\ref{eq:attention_weights}) and Eq~(\ref{eq:attention_output}). 
The output token $\mathbf Y_t$ is subsequently concatenated with previously generated tokens to serve as input for the next step in the token generation sequence.

\noindent \textbf{Contrastive Decoding for LVLMs}.
Contrastive Decoding~\cite{li2022contrastive} is an effective training-free intervention strategy for suppressing hallucinations in LVLMs. Its core idea is to introduce a manually crafted, distorted input to induce the model to generate hallucinated outputs, which are then used to mitigate potential hallucinations in the original output by contrasting original output distribution logits to the distorted counterpart. Existing works propose designing distorted inputs—either textual~\cite{icd} or visual~\cite{leng2024mitigating}—to counteract \textit{hallucinations caused by uni-modal priors}, where the model generates results based solely on one modality while ignoring the other. Formally, contrastive decoding can be expressed as:
\begin{align}
& \mathbf l_t =  \textrm{logit}_{\theta}(\mathbf Y_t|\textrm{Attn}(\mathbf O | \mathbf T, \mathbf X, \mathbf Y_{<t})), \\
& \widetilde{\mathbf l}_t = \textrm{logit}_{\theta}(\mathbf Y_t|\textrm{Attn}(\mathbf O |\mathbf{\widetilde{T}}, \mathbf{\widetilde{X}}, \mathbf Y_{<t})), \\
 & p_{\theta}(\mathbf Y_t|\mathbf T, \mathbf X) =  \textrm{softmax}((1 + \alpha) \mathbf l_t   -  \alpha \widetilde{\mathbf l}_t).
\label{eq:contrastive}
\end{align}
$\alpha$ is the hyper-parameter controlling the contrastive effect,$\mathbf{\widetilde{T}}$ and $\mathbf{\widetilde{X}}$ are the distorted text and visual input, respectively. Despite widely studied~\cite{leng2024mitigating,chen2024halc,woo2024don,huo2024self,wang2024mitigating}, existing methods primarily focus on the unimodal distortion, neglect the spurious inter-modality correlations, leading to suboptimal estimation of distorted distribution.

\subsection{IMCCD for LVLMs Hallucination Mitigation}

\begin{figure*}[t]
\centering

\includegraphics[width=0.99\textwidth]{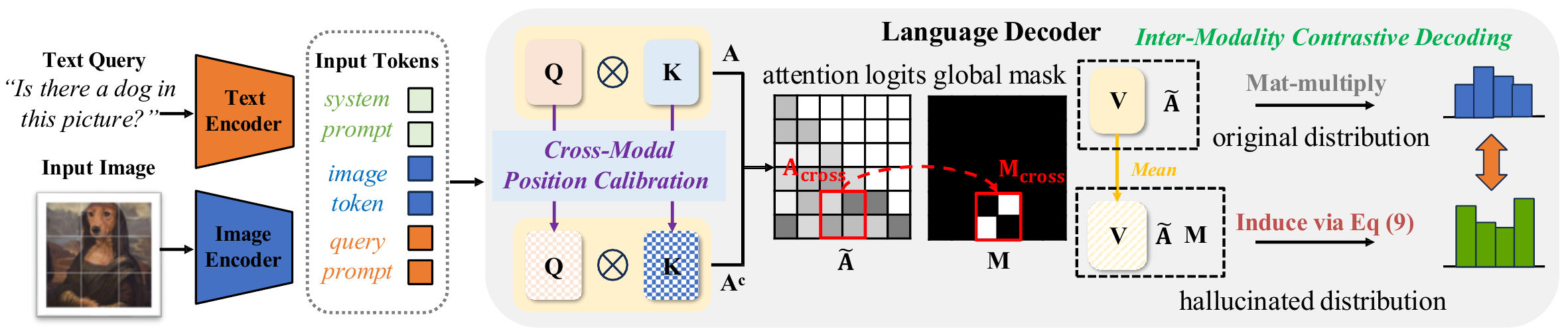}
\caption{An overview of the proposed CMAC approach, consisting of two modules: Inter-Modality Decoding (IMD)  and Cross-Modal Position Calibration (CMPC). During inference, IMD generates a distorted output distribution that favors hallucination by selectively masking value vectors corresponding to high attention weights in the cross-modal segment of the attention matrix. 
IMD then performs contrastive decoding between the original outputs and distorted outputs to mitigate the hallucination. Additionally, CMPC scales the position indexes to estimate the content-driven attention weights to replace the cross-modal segment, alleviating the position bias.
% refines the cross-modal segments of attention logits with content-driven attention logits estimated by normalizing the position indices of all image tokens to a uniform value.
% BCP is first proposed to discover and represent background underlying categories, estimated from background proposals, with learnable background category-specific contexts. Then, BOD is presented to employ $k$-means clustering on background proposals across all images to harness implicit objects explored from background underlying knowledge. During inference, IPR is introduced to rectify probability scores of novel categories provided, by loosening their conceptual overlaps with background underlying categories estimated from background proposals.
% In the training stage, BCP represents background with learnable background category-specific context . BOD identifies implicit objects correlated with underlying categories in background proposals by performing K-means clustering on  background proposals on all images. In inference stage, IPR rectifies the probability scores of novel class to address conceptual overlaps between background underlying categories and novel classes.
}%\vspace{-3mm}
\label{fig:pipeline}
% \vspace{-0.50cm}
\end{figure*}

In this section, we present the Cross-Modal Attention Calibration (CMAC) framework, designed to mitigate hallucinations in LVLMs. CMAC incorporates two key components: Inter-Modality Decoding (IMD) and Cross-Modal Position Calibration (CMPC), which are used to address the hallucination caused by the spurious inter-modality correlations and positional biases, respectively. An overview of our framework is illustrated in Figure~\ref{fig:pipeline}.

\subsubsection{Inter-Modality Contrastive Decoding}
% Existing contrastive decoding methods mainly distort the visual content to counteract the hallucination from language priors. Nevertheless, they neglect the hallucination from spurious inter-modality correlations. To mitigate this issue, we propose Cross-Modal Value-Enhanced Decoding(IMD), which performs distortion on cross-modal attention by selectively masking the value vectors corresponding to high values in attention weights.

% Existing contrastive decoding methods primarily distort visual content to counteract hallucinations driven by overreliance on language priors. However, these approaches overlook hallucinations stemming from spurious inter-modality correlations. To address this limitation, we propose Cross-Modal Value-Enhanced Decoding (IMD), which performs distortion in cross-modal attention by selectively masking the value vectors associated with high attention weights. 

% important value vectors with high attention weights in the cross-modal part of the self-attention layer. 
Existing contrastive decoding methods primarily address the hallucinations caused by over-reliance on the uni-modal priors(e.g., over-reliance on either text or vision) through input distortions, while neglecting \textit{hallucinations arising from spurious inter-modality correlations}. To bridge this gap, we propose Inter-Modality Decoding (IMD), which corrects cross-modal correlations through targeted distortion of cross-modal attention patterns. Our key insight is to selectively suppress the value vectors corresponding to high values in attention logits and exacerbate the hallucination from the unreliable inter-modality interaction, thus subsequently alleviating it with contrastive decoding.

% Existing contrastive decoding methods primarily focus on the hallucination stemming from the over-reliance on the uni-modal prior by distorting the textual or visual content, ignoring the hallucination caused by the spurious inter-modality correlation. To this end, we propose inter-modality decoding(IMD), which performs inter-modality correlation correction via the distortion on the level of cross-modal attention.

% Our key insight is to selectively suppress the value vectors corresponding to high values in attention logits and exacerbate the hallucination from the inter-modality interaction, thus subsequently alleviating it with contrastive decoding. 
% Concretely, for each head in each attention layer of the language decoder, we define the attention logits before the softmax function in Eq.~(\ref{eq:attention_weights}) as,
% \begin{equation}
% \mathbf{A}^l = \frac{\mathbf{QK}^T}{\sqrt{d}} . 
% \label{eq:attention_logits}
% \end{equation}
Specifically, we firstly isolate the cross-modal segment from the attention logs $\mathbf{A}^l \in \mathbb R^{(n+m)\times (n+m)}$, which is denoted as $\mathbf{A}^{l}_{\textrm{cross}} = \mathbf{A}^{l}[m_b+n: m+n, m_b:m_b+n]$. Generally, $\mathbf{A}^{l}_{\textrm{cross}}$ represents the attention weights between the subsequent text tokens and image tokens, capturing the inter-modality correlations within the self-attention layer.
% However, the attention weights fail to reflect the true inter-modality correlations due to the position bias. To mitigate the position bias, IMD first estimates a position-based query $\mathbf Q_{\textrm b})$ by replacing the value of text query $Q[m_b+n:m+n]$ to their mean value $\mu_{t}(Q[m_b+n:m+n])$ and applying ROPE on the refined text queries. $\mu_{t}$ denotes the token-wise mean operation. The values in $\mathbf Q_{\textrm b})$ do not reflect the meaning of text tokens while influenced by the position embeddings.
% Then IMD estimates the bias map by performing attention bewteen the  $\mu(Q_{\textrm b})$ and the image keys $\mathbf K_{\textrm p}[m_b:m_b+n]$,
% \begin{equation}
% \mathbf B = \frac{\mathbf{Q_{\textrm b}K_{\textrm  p}[m_b:m_b+n]}^T}{\sqrt{d}}. \\
% \label{eq:mask}
% \end{equation}
Based on such cross-modal attention weight, we aim to estimate the distribution that favors the hallucination, which afterward is used to mitigate the hallucination from spurious inter-modality correlations. To accomplish this, IMD first generates a mask that selectively identifies the prominent attention weights based on their magnitude,
\begin{equation}
\mathbf M_{\textrm{cross}} =  \mathbb{I}(\mathbf{A}^{l}_{\textrm{cross}} >  \mu(\mathbf{A}^{l}_{\textrm{cross}})), \\
\label{eq:mask}
\end{equation}
where  $\mu(\cdot)$ represents the mean operation and $\mathbb{I}$ is the indicator function. 
The binary mask $\mathbf M_{\textrm{cross}}  \in \mathbb R^{(m - m_b)\times (n)}$  adaptively indicates the selected significant attention weights without the position bias for the cross-modal segment. We subsequently pad the $\mathbf M_{\textrm{cross}}$ to a global mask $\mathbf M \in \mathbb R^{(n+m)\times (n+m)}$ which applies to all attention weights in $\mathbf{A}$ with zero padding. Specifically, $\mathbf{M}[m_b+n: m+n, m_b:m_b+n]$ is set to the $\mathbf M_{\textrm{cross}}$, ensuring that the cross-modal part is accurately masked in the global context.

% To estimate the distribution favored hallucination, we aim to design a new forward process $\textrm{logits}^s_{\theta}$ for language decoder to distort the significant part in $\mathbf{A}^{l}_{cross}$ when keeping the remainder in $\mathbf A^{l}$ unchanged to mitigate the hallucination from spurious inter-modality correlations.
% To accomplish this, IMD first generates a mask that selectively identifies the prominent attention weights based on their magnitude,
% \begin{equation}
% \mathbf M_{cross} =  \mathbb{I}(\mathbf{A}^{l}_{cross} <  \mu(\mathbf{A}^{l}_{cross})), \\
% \label{eq:mask}
% \end{equation}
% where $ \mu(\cdot)$ represents the mean operation and $\mathbb{I}$ is the indicator function. The binary mask $\mathbf M_{cross}  \in \mathbb R^{(m - m_b)\times (n)}$  adaptively indicates the selected significant attention weights for the cross-modal segment. We subsequently pad the $\mathbf M_{cross}$ to a global mask $\mathbf M \in \mathbb R^{(n+m)\times (n+m)}$ which applies to all attention weights in $\mathbf{A}$ with zero padding. Specifically, $\mathbf{M}[m_b+n: m+n, m_b:m_b+n]$ is set to the $\mathbf M_{cross}$, ensuring that the cross-modal part is accurately masked in the global context.

With this global mask, IMD then distorts the cross-modal segment of self-attention. While recent approaches typically adapt token-wise pruning image tokens, IMD masks the value vector $\mathbf V$ to $\mu(V)$ by dim-wise mean operation to exacerbate the hallucination. With this process, the content of the value vector is teased apart.
% \begin{equation}
% \widetilde{\mathbf V} =  
% \label{eq:mask}
% \end{equation}
% With the global mask, IMD then performs the distortion on the cross-modal part of self-attention. Recent studies distort the visual part by directly adding noise to image pixel values or pruning out image tokens. Differently, we propose to distort the cross-modal attention by adding noise to the value vectors. Following the diffusion process in VCD, the distorted value vectors ${\mathbf V}_{S}$ can be formulated as the following equations,
% \begin{align}
% &\textrm q\left({\mathbf V}_t \mid {\mathbf V}_{t-1}\right)=\mathcal{N}\left({\mathbf V}_t ; \sqrt{1-\gamma} {\mathbf V}_{t-1}, \gamma \mathbf{I'}\right) \\
% &\textrm q\left({\mathbf V}_{S} \mid {\mathbf V}\right)=\prod_{t=1}^S \textrm q\left({\mathbf V}_t \mid {\mathbf V}_{t-1}\right).
% \end{align}
% $S$ is the step of the diffusion process.
IMD then estimates the attention output by performing the weighted sum between the origin value vector $\mathbf V$ and distorted value vector $\mu(\mathbf V)$  guided by both the mask $\mathbf{M}$ and attention weights $\mathbf{A}$. Specifically, IMD modifies the process Eq~(\ref{eq:attention_output}) by the following process, 
\begin{equation}
\widetilde{\mathbf O} = (\mathbf M \cdot \mathbf A) \mu(\mathbf V) + ((1 - \mathbf M) \cdot \mathbf A) \mathbf V.
\label{eq:distorted_attention}
\end{equation}
With the above equation, IMD distorts the value vectors corresponding to the significant cross-modal attention weights while keeping the value vectors of the remaining attention part unchanged, which suppresses important inter-modality correlations from cross-modal attention.
Similar to Eq~(\ref{eq:contrastive}), the final contrastive decoding process is performed by the following equations,
\begin{align}
% p_{\theta}(\mathbf Y_t|\mathbf T, \mathbf X) = & \textrm{softmax}[(1 + \alpha)\textrm{logit}_{\theta}(\mathbf Y_t|\mathbf T, \mathbf X, \mathbf Y_{<t}) \nonumber \\
% & -  \alpha \textrm{logit}^{s}_{\theta}(\mathbf Y_t|\mathbf{T}, \mathbf{X}, \mathbf Y_{<t})].
& \mathbf l_t =  \textrm{logit}_{\theta}(\mathbf Y_t|\textrm{Attn}(\mathbf O | \mathbf T, \mathbf X, \mathbf Y_{<t})), \\
& \widetilde{\mathbf l}_t = \textrm{logit}_{\theta}(\mathbf Y_t|\textrm{Attn}(\widetilde{\mathbf O}|\mathbf T, \mathbf X, \mathbf Y_{<t})), \\
 & p_{\theta}(\mathbf Y_t|\mathbf T, \mathbf X) =  \textrm{softmax}((1 + \alpha) \mathbf l_t   -  \alpha \widetilde{\mathbf l}_t).
\label{eq:contrastive_IMD}
\end{align}

Compared with recent contrastive decoding methods, the advantages of IMD are threefold: 
(1) The distortion process used in IMD directly focuses on the cross-modal segment of attention to alleviate the cross-modal inconsistency. Specifically, it proposes to modify the value vectors with high attention weights on cross-modal attention, leaving the low-weight inter-modality correlations and intra-modality knowledge exchange unaffected. This approach enables IMD to address hallucinations from both uni-modality overreliance and spurious inter-modality correlations more effectively.
(2) Traditional contrastive decoding methods alter attention weights from visual tokens to text tokens, which can result in over- or under-estimated hallucinations from image regions. IMD, however, avoids modifying attention weights, yielding a more precise estimation of distributions prone to hallucination.
(3) Besides, since the attention process of the visual part remains the same, the key and value vectors and attention weights can be directly derived from the original forward process, leading to a faster inference time than other contrastive decoding methods.

\subsubsection{Cross-Modal Position Calibration}
Although IMD reduces hallucinations arising from the cross-modal interaction, MLLMs still face challenges with visual content retention. Specifically, with the position embeddings, the text tokens will be biased to pay more attention to the latter part of the image visual tokens, while overlooking the head part of the visual tokens,  leading to more hallucinations. 
Motivated by this, we propose a cross-modal position calibration module to encourage the self-attention layer to alleviate the position bias from position embeddings on the cross-modal part. Concretely,  we estimate new position indexes by scaling the position indices of all image tokens.
It is achieved by replacing the original position indexes of concatenated input tokens, noted with $\mathbf P = [\{i\}_{i=1}^{m+n}]$, with refined position indexes,
\begin{equation}
\mathbf P^{c} = [\{i\}_{i=1}^{m_b}, \{m_b+  \frac{i}{\gamma}\}_{i=1}^{n}, \{i+\frac{n}{\gamma}\}_{i=m_b+1}^{m}],
\end{equation}
where $\gamma$ is the scaling factor.

The refined position indexes shrink the gap between rotary angles of image tokens in rotary matrices in Eq~ (\ref{eq:rope}).
This keeps the global positional relation of the image content in the input language and reduces the positional bias of the head part of image tokens with other language tokens. 
The modified position indexes $\mathbf{P}^{c}$ are used to perform ROPE  to estimate the attention logits $\mathbf A^c$ with refined position embeddings by Eq~ (\ref{eq:rope}) and Eq~(\ref{eq:attention_weights}).
Then we refine the cross-modal part of attention logits $\mathbf A^{l}$ with,
\begin{equation}
\label{eq:refinement}
\widetilde{\mathbf{A}}^{l}_{i,j}  =
\begin{cases}
\mathbf{A}^{c}_{i,j},  & \text{if} \; j > m^b+n \;
 \text{and} \; m^b+n > i \geq m^b , \nonumber \\
\mathbf{A}^{l}_{i,j}, & \text{otherwise}.
\end{cases}
\end{equation}
The $\widetilde{\mathbf{A}}^{l}$ is used to perform the further estimation of values in Eq~(\ref{eq:mask}), Eq~(\ref{eq:attention_output}) and Eq~(\ref{eq:distorted_attention}).
Note that this refinement is applied to both the estimation of logits from original inputs and distorted inputs. 
% Specifically, the $\mathbf{A}^{l}_{\textrm{cross}}$ in Eq.~\ref{eq:mask} and $\mathbf{A}$ in Eq.~\ref{eq:attention_output}, Eq.~\ref{eq:distorted_attention} are replaced with $\widetilde{\mathbf{A}}^{l}_{\textrm{cross}}$ and $\widetilde{\mathbf{A}}$, respectively. $\widetilde{\mathbf{A}}^{l}_{\textrm{cross}}$ is the cross-modal segment of $\widetilde{\mathbf{A}}^{l}$ and $\widetilde{\mathbf{A}} = \textrm{softmax}(\widetilde{\mathbf{A}}^{l})$ is the refined attention weights.
% Specifically, we replace the Eq.~\ref{eq:mask}, Eq.~\ref{eq:attention_output}, Eq.~\ref{eq:distorted_attention} with.
% \begin{align}
% &\mathbf M_{cross} =  \mathbb{I}(\widetilde{\mathbf{A}}^{l}_{cross} <  \mu(\widetilde{\mathbf{A}}^{l}_{cross})), \\
% &\mathbf O =  \widetilde{\mathbf{A}}\mathbf{V}, \\
% & \widetilde{\mathbf O} = (\mathbf M \cdot \widetilde{\mathbf{A}}) \mu(\mathbf V) + ((1 - \mathbf M) \cdot \widetilde{\mathbf{A}}) \mathbf V.
% \end{align}
With CMPC, the language decoder is encouraged to prioritize image content over token positions, effectively alleviating hallucinations caused by the overlook of visual content. We provide some further analysis about the position bias of RoPE in the supplementary document.

\section{Experiments}
% \subsection{Datasets and Evaluation Metrics}
% % Following VCD~\cite{leng2024mitigating}, we evaluate our proposed method on benchmark datasets, POPE, and MME datasets for the hallucination of LVLMs. We also report the results on CHAIR metrics on the MS-COCO subset for the image captioning task.
% We evaluate our method on the widely used benchmarks of LVLMs hallucination mitigation, including POPE, CHAIR, and MME. POPE is a VQA dataset for object hallucination, which consists of nine different subsets from three datasets, MS-COCO, AOKVQA, and GQA with three different negative sampling strategies (Random, Popular, and Adversarial).  The CHAIR benchmark is used to assess the hallucination on the image captioning task.
% It set up two metrics, namely the $\text{CHAIR}_i$ and $\text{CHAIR}_s$, which are formulated as $\text{CHAIR}_i = \frac{|\{\text{hallucinated objects}\}|}{|\{\text{all objects mentioned}\}|}$, $\text{CHAIR}_s = \frac{|\{\text{sentences with hallucinated object}\}|}{|\{\text{all sentences}\}|}.$ Following previous work\cite{huang2024opera},  we randomly sample 500 images in the validation set of MS-COCO and conduct the evaluation of CHAIR metrics on the sampled subset for the image captioning task with the prompt "Please describe this image in detail". The MME dataset is a comprehensive VQA benchmark for the evaluation of LVLMs. For hallucination evaluation, we report the results on both the MME full set and the hallucination subset with four hallucination-related sub-tasks. More details about datasets are shown in our supplementary materials.

\subsection{Datasets and Evaluation Metrics}
\noindent \textbf{POPE}. POPE~\cite{li2023evaluating} designs a new metric and benchmark to assess object hallucination in the VQA paradigm. It achieves the evaluation of object hallucination in a binary classification task by prompting LVLMs to answer yes or no for short questions about the existence of probing objects (e.g., Is there a car in the image?).  The POPE benchmark consists of nine different subsets from three datasets, MS-COCO, AOKVQA, and GQA, with three different negative sampling strategies (Random, Popular, and Adversarial). Each sampled subset includes 500 images with 6 problems for each image. Following previous works~\cite{leng2024curse,wang2024mitigating}, we add a " Please answer in one word." constraint for evaluating the POPE dataset at the end of the query text. F1-score and Accuracy are commonly reported as the metrics for the performance of hallucination mitigation on this benchmark.

\noindent \textbf{Chair}. The CHAIR~\cite{rohrbach2018object} is a widely used metric for assessing the hallucination in responses of LVLMs. The CHAIR metric comprises two variants, namely the $\text{CHAIR}_i$ and $\text{CHAIR}_s$, which are formulated as follows:
\begin{equation*}
    \text{CHAIR}_i = \frac{|\{\text{hallucinated objects}\}|}{|\{\text{all objects mentioned}\}|},
\end{equation*}
\begin{equation*}
    \text{CHAIR}_s = \frac{|\{\text{sentences with hallucinated object}\}|}{|\{\text{all sentences}\}|}.
\end{equation*}
Besides, we also report the recall and F1-score to show the completeness of the generated caption of the image. Following previous work\cite{huang2024opera},  we randomly sample 500 images in the validation set of MS-COCO and conduct the evaluation of CHAIR metrics on the sampled subset with the prompt "Please describe this image in detail". 
%We report the results under max new token numbers of 1024. 

\noindent \textbf{MME}. The MME dataset is a comprehensive benchmark for the evaluation of LVLMs. 
Similar to POPE, these tasks are formulated as binary classifications with yes-no questions. It encompasses 14 subtasks for the examination of the perception and cognition abilities of LVLMs. The score is estimated by the sum of the accuracy of each question and the accuracy of each image.
For hallucination evaluation, we report the results on both the MME full set and the hallucination subset with four hallucination-related sub-tasks, including object existence, count, position, and color.

\subsection{Models and Implementation Details}
We follow VCD\cite{leng2024mitigating} to integrate our proposed method with three popular LVLMs, LLaVA1.5~\cite{liu2024improved}, InstructBLIP~\cite{Instructblip} and QWEN-VL~\cite{bai2023qwenvl}. All these models adopt Vicuna 7B~\cite{vicuna2023} as their language models. 
We empirically set  $\gamma = 2$ and $\alpha=3$. On all the sampling setting, top p is set to 1. We leave other hyper-parameters the same with VCD.

\subsection{Experimental Results}
\noindent\textbf{Comparison on POPE.}
We compare our method against three existing contrastive decoding methods, including VCD~\cite{leng2024mitigating}, ICD~\cite{wang2024mitigating}, OPERA~\cite{huang2024opera} and PAI~\cite{liu2024paying}, on the POPE dataset. In the POPE dataset, we report the average of performance on COCO, AOKVQA, and GQA. We also follow VCD to report the performance of different subsets of POPE distinctly, which is shown in our supplementary document.
The results are presented in Table~\ref{tab:pope_average}.
It can be observed that our method boosts the performance of LVLMs in mitigating hallucination from object existence for each setup of different subsets on POPE. Our proposed CMAC method outperforms the existing methods by a clear margin on different LVLMs. 
Compared to the second-best method PAI, CMAC achieves a notable increase in accuracy and F1 score, underscoring its effectiveness in reducing spurious inter-modality correlations. For greedy search, the LVLMs are more sensitive to distribution so some methods may leads to performance degradation. Nevertheless, the results showcases our CMAC also can achieve significantly better results with greedy search,  which demonstrates the generalization capability of CMAC. 

% Compared with VCD, the CMAC consistently leads to an significant improvement on both accuracy (from 0.76\% to 3.27\%) and F1 score (from 0.53\% to 2.57\%). These suggest the enhanced capability of CMAC in alleviating spurious inter-modality correlations.
% Specifically, since previous methods may over- or under- estimates the hallucination by altering the attention weights from visual contents with distortion, they always 
% Besides, CMAC leads to performance improvement on both precision and recall on mosts of setups, which demonstrates the generalization capability of CMAC to generate both precise and comprehensive responses.

\begin{table*}[t] 
\centering
\small
\begin{tabular}{ll|cc|cc|cc||cc|cc|cc}
\toprule
% --- Header Row 1: Decoding Strategy ---
& & \multicolumn{6}{c}{\textbf{Sampling}} & \multicolumn{6}{c}{\textbf{Greedy}} 

\\
\cmidrule(lr){3-8} \cmidrule(lr){9-14}

% --- Header Row 2: Setting ---
& & \multicolumn{2}{c|}{\textbf{Random}} & \multicolumn{2}{c|}{\textbf{Popular}} & \multicolumn{2}{c||}{\textbf{Adversarial}} & \multicolumn{2}{c|}{\textbf{Random}} & \multicolumn{2}{c|}{\textbf{Popular}} & \multicolumn{2}{c}{\textbf{Adversarial}} \\
\cmidrule(lr){3-4} \cmidrule(lr){5-6} \cmidrule(lr){7-8} \cmidrule(lr){9-10} \cmidrule(lr){11-12} \cmidrule(lr){13-14}

% --- Header Row 3: Metrics ---
\textbf{} & \textbf{Method} & Acc & F1  & Acc & F1  & Acc & F1 & Acc & F1  & Acc & F1  & Acc & F1 \\ 
\hline \hline

% --- LLaVA-1.5 Data ---
\multirow{6}{*}{\rotatebox{90}{\textbf{LLaVA-1.5}}} 
& Baseline & 83.49 & 82.28 & 79.98 & 79.34 & 76.03 & 76.26 & 83.51 & 82.27 & 79.68 & 79.07 & 75.91 & 76.11 \\
& ICD    & 85.70 & 84.43 & 81.61 & 81.46 & 78.66 & 78.57 & 87.82 & 86.73 & 84.72 & 83.94 & 80.98 & 80.78 \\
& VCD      & 86.84 & 86.83 & 82.65 & 83.37 & 77.31 & 79.28 & 88.11 & 88.15 & 84.23 & 84.91 & 78.08 & 80.07 \\
& PAI      & 87.73 & 87.65 & 83.45 & 84.21 & 78.36 & 78.53 & 88.32 & 88.01 & 84.69 & 85.61 & 79.18 & 80.38 \\
& OPERA          & -     & -     & -     & -     & -     & -     & 88.85 & 88.67 & 82.77 & 83.40 & 79.16 & 80.93 \\
& \textbf{Ours}  & \textbf{89.10} & \textbf{88.60} & \textbf{86.0} & \textbf{85.70} & \textbf{81.41} & \textbf{81.84} & \textbf{89.33} & \textbf{88.89} & \textbf{86.33} & \textbf{86.00} & \textbf{82.02} & \textbf{82.42} \\
\hline

% --- InstructBLIP Data ---
\multirow{6}{*}{\rotatebox{90}{\textbf{InstructBLIP}}} 
& Baseline & 80.41 & 80.94 & 76.09 & 77.65 & 72.37 & 75.42 & 87.44 & 86.80 & 81.85 & 82.58 & 77.66 & 79.21 \\
& ICD     & 85.78 & 85.73 & 81.12 & 82.25 & 76.82 & 78.99 & 86.76 & 86.83 & 80.54 & 82.16 & 76.73 & 78.83 \\
& VCD      & 84.11 & 84.13 & 79.96 & 80.80 & 76.32 & 78.08 & 86.66 & 86.37 & 81.73 & 82.29 & 78.00 & 79.45 \\
& PAI      & 84.60 & 84.54 & 80.13 & 80.94 & 77.00 & 78.33 & 86.51 & 86.19 & 81.65 & 82.34 & 77.76 & 79.13 \\
& OPERA          & -     & -     & -     & -     & -     & -     & 84.57 & 83.74 & 78.24 & 79.15 & 74.59 & 76.33 \\
& \textbf{Ours}& \textbf{86.67} & \textbf{86.10} & \textbf{82.20} & \textbf{82.37} & \textbf{78.37} & \textbf{79.25} & \textbf{87.68} & \textbf{87.36} & \textbf{82.38} & \textbf{82.80} & \textbf{78.77} & \textbf{79.58} \\
\hline

% --- QWEN-VL Data ---
\multirow{6}{*}{\rotatebox{90}{\textbf{Qwen-VL}}} 
& Baseline & 84.12 & 82.42 & 81.89 & 80.51 & 79.10 & 78.07 & 84.09 & 82.77 & 83.97 & 82.63 & 81.40 & 80.38 \\
& ICD     & 86.45 & 85.80 & 83.41 & 83.27 & 81.04 & 81.33 & 86.49 & 85.86 & 83.80 & 83.71 & 80.96 & 81.27 \\
& VCD      & 87.82 & 87.68 & 85.60 & 85.48 & 81.85 & 82.34 & 87.83 & 87.39 & 85.91 & 85.78 & 81.90 & 82.44 \\
& PAI     & 88.15 & 87.94 & 85.73 & 85.51 & 81.98 & 82.23 & 87.86 & 87.63 & 85.73 & 86.62 & 82.06 & 82.44 \\
& OPERA          & -     & -     & -     & -     & -     & -     & 83.05 & 83.20 & 81.40 & 81.89 & 76.73 & 78.31 \\
& \textbf{Ours}  & \textbf{88.61} & \textbf{88.26} & \textbf{86.67} & \textbf{86.34} & \textbf{82.68} & \textbf{83.05} & \textbf{88.92} & \textbf{88.48} & \textbf{86.92} & \textbf{86.55} & \textbf{83.01} & \textbf{83.47} \\
\bottomrule
\end{tabular}
\caption{\textbf{Results on discrimination hallucination benchmark POPE.} The Baseline method denotes the standard decoding. The best performances within each setting are \textbf{bolded}. "Acc" and "F1" denote the Accuracy and F1 scores, respectively. The results are copied from the papers or from our re-implementation based on official codes. }
\label{tab:pope_average}
\end{table*}

\noindent\textbf{Comparison on CHAIR.}
We conduct evaluation of the CHAIR metric on the MS-COCO validation set in Table~\ref{tab:chair} to validate the performance of our method on the image captioning task and long sequence generation. 
The table shows that the performance of our method is better than that of other existing contrastive decoding methods. Specifically, our CMAC significantly enhances the performance of LVLMs on $\text{CHAIR}_i$ and $\text{CHAIR}_s$ by 9.6\% and 5.1\% for LLaVA 1.5 and by 6.4\% and 3.5\% for InstructBLIP under nucleus sampling, demonstrating the capability of our method for generating long sequence response with less hallucination. 
Besides, our CMAC also achieves a better F1 score, which showcases that the generated responses from our method can describe the image content more accurately and completely.
\begin{table*}[]
\centering
\small
\begin{tabular}{ccccccccccccc}
\toprule
\multirow{2}{*}{
\textbf{Method}} & \multicolumn{4}{c}{\textbf{LLaVA 1.5}} & \multicolumn{4}{c}{\textbf{InstructBLIP}}  & \multicolumn{4}{c}{\textbf{Qwen-VL}}\\ \cmidrule(lr){2-5} \cmidrule(lr){6-9} \cmidrule(lr){10-13} 
  &   $\textbf{C}_i \downarrow$    &    $\textbf{C}_s\downarrow$   &  \textbf{Recall} $\uparrow$    &     \textbf{F1} $\uparrow$ &        $\textbf{C}_i \downarrow$    &    $\textbf{C}_s\downarrow$   &  \textbf{Recall} $\uparrow$    &     \textbf{F1} $\uparrow$ &    $\textbf{C}_i \downarrow$    &    $\textbf{C}_s\downarrow$   &  \textbf{Recall} $\uparrow$    &     \textbf{F1} $\uparrow$     \\ \hline
Sampling &    55.6   &   17.8    &   72.4    &   77.0    &  57.0      &  16.7      &    70.4    &    76.3  & 44.8      &  11.3     &    74.6    &  81.1 \\
 VCD &    54.2   &   16.4    &   76.7    &     80.0  &   53.6    &   15.6     &  \textbf{74.6}      &    79.2  &  47.2     &  12.1      &    \textbf{76.1}    &    81.6 \\
 ICD &    54.0   &    15.4   &   \textbf{77.2}    &   80.7    &    \textbf{49.6}    &    14.5    & 72.6    &   78.5   &  43.2      &  11.2      &    75.0    &    81.5 \\
 Ours  &   \textbf{47.0}    &   \textbf{12.7}    &   75.6    &  \textbf{81.0}     &   50.6      &   \textbf{13.2}     &   73.7     & \textbf{79.7}  &  \textbf{41.2}      &  \textbf{10.6}      &    75.4    &    \textbf{81.8}   \\ 

\cmidrule{1-13}
Greedy &    49.6   &   15.1    &   77.9    &   81.2    &  48.2      &  13.3      &    73.8    &    79.7  &  45.8      &  11.4      &    74.6   &    81.0 \\
 VCD &    55.2   &   15.7    &   77.6    &     80.8  &   49.8   &   14.0  &  \textbf{74.9}      &    80.1 &  46.6      &  12.2     &    \textbf{76.3}    &    81.7 \\
 ICD &    57.0   &    15.4   &   \textbf{77.5}    &   80.9    &    \textbf{44.6}    &    12.6    & 73.9   &  80.1   &  42.6     &  11.1      &    74.9   &    81.3 \\
 Ours  &  \textbf{47.6}    &   \textbf{12.5}    &   76.2    &  \textbf{81.5}     &   45.2      &   \textbf{12.1}     &   74.2    & \textbf{80.5}  &  \textbf{41.4}     &  \textbf{10.9}      &    75.6    &    \textbf{81.8}  \\ \bottomrule
\end{tabular}
\caption{\textbf{Results on hallucination metric CHAIR on the MS-COCO validation set.}  The best performances within each setting are \textbf{bolded}. $\textbf{C}_i$ and  $\textbf{C}_s$  denotes the $\text{CHAIR}_i$ and $\text{CHAIR}_s$, respectively.}
\label{tab:chair}
\end{table*}

\noindent\textbf{Comparison on MME.} To validate the effectiveness of our method in mitigating various types of hallucination beyond object existence, we perform the comparison on the MME hallucination subset and the MME full set with nucleus sampling. The results are averaged over 4 runs.
For MME hallucination set shown in Table~\ref{tab:mme}, 
CMAC achieves better scores on mitigating both object-level and attribute-level hallucination. Beyond the object existence, CMAC leads to an improvement of 17.50, 8.50, and 5.42 on count, position, and color for LLaVA1.5, which demonstrates the generalization capability of CMAC across various types of hallucinations. 

\begin{figure}[h]
\centering
% \vspace{-0.25cm}
% \includegraphics[width=\textwidth, trim=0 0 0 5, clip]{images/pipeline_final.pdf}
    \setlength{\abovecaptionskip}{0.cm}
\includegraphics[width=0.97\linewidth, trim=0 0 0 0, clip]{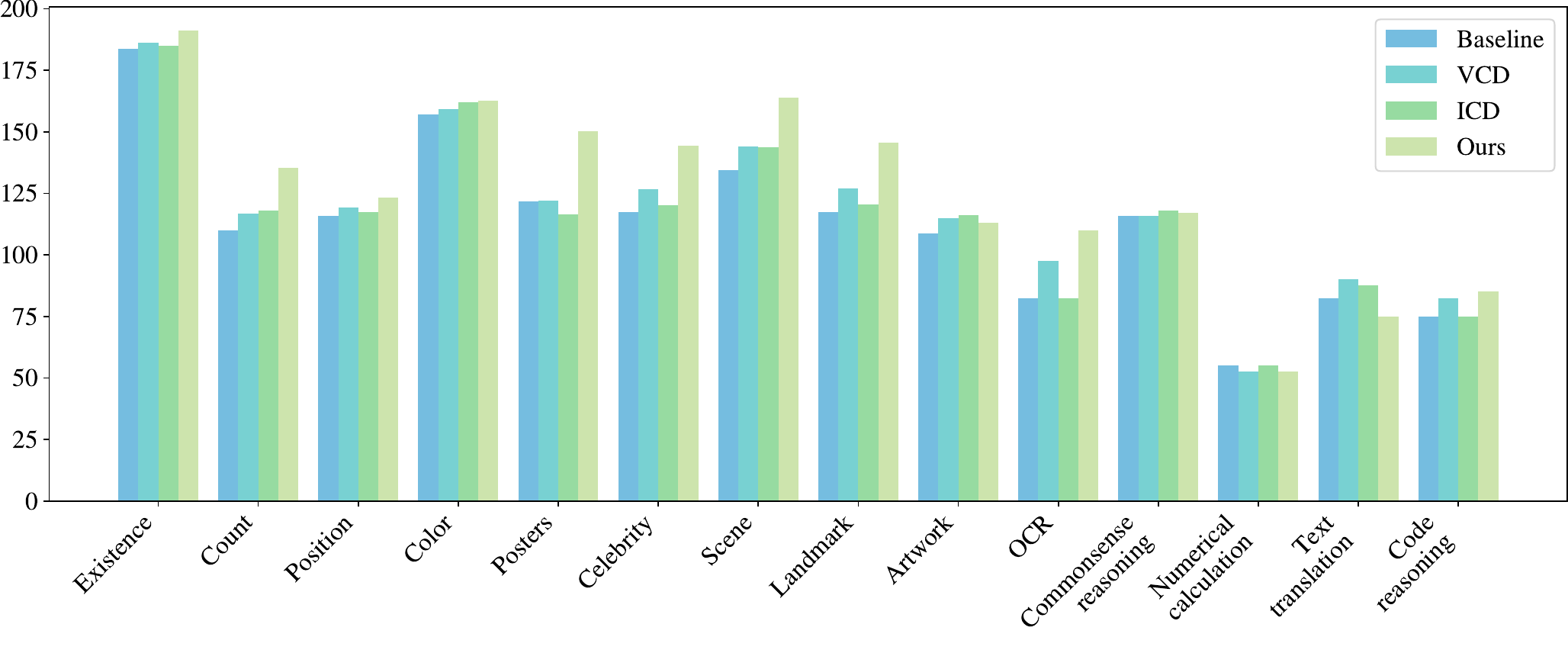}
% \vspace{-3mm}
% \vspace{-0.25cm}
\caption{ Results of LLaVA1.5 on MME-Fullset. 
}%\vspace{-3mm}
\label{fig:mme_full}
% \vspace{-0.50cm}
\end{figure}

\begin{table}[h]
\centering
\setlength\tabcolsep{3pt}
\fontsize{7.5}{9}\selectfont
\begin{tabular}{@{}ccccccc@{}}
\toprule
\multirow{2}{*}{\textbf{Model}}        & \multirow{2}{*}{\textbf{Decoding}} & \multicolumn{2}{c}{\textbf{Object-level}}                                   & \multicolumn{2}{c}{\textbf{Attribute-level}}                               & \multicolumn{1}{c}{\multirow{2}{*}{\textbf{Total}$\uparrow$}} \\
                              &                           & \multicolumn{1}{c}{\textit{Existence}$\uparrow$} & \multicolumn{1}{c}{\textit{Count}$\uparrow$} & \multicolumn{1}{c}{\textit{Position}$\uparrow$} & \multicolumn{1}{c}{\textit{Color}$\uparrow$} & \multicolumn{1}{c}{}                       \\ \midrule
\multirow{4}{*}{LLaVA1.5}     & Baseline                 &$183.75$ &$110.00$ &$115.83$ &$157.08$ &$566.67$ \\
                              & VCD                       &$186.25$ &$116.67$ &$119.17$ &$159.17$ &$581.25$ \\
                              & ICD                       &$185.00$ &$117.91$ &$117.50$ &$162.08$ &$582.50$    \\        
                              & Ours                      &$\textbf{191.25}$ &$\textbf{135.41}$ &$\textbf{123.33}$  &$\textbf{162.50}$ &$\textbf{612.49}$    \\       
                               \midrule
\multirow{4}{*}{InstructBLIP}     & Baseline                 &$154.17$ &$\textbf{86.67}$ &$58.33$ &$123.75$ &$422.91$ \\
                              & VCD                       &$166.67$ &$82.08$ &$67.08$ &$130.00$ &$445.83$ \\
                              & ICD                       &$166.67$ &$80.83$ &$67.92$ &$128.33$ &$443.75$    \\        
                              & Ours                      &$\textbf{183.75}$ &$ 85.00$ &$\textbf{68.75}$ &$\textbf{131.67}$ &$\textbf{469.17}$    \\ 
                              \midrule
\multirow{4}{*}{Qwen-VL}     & Baseline                 &$142.50$ &$116.67$ &$134.08$ &$157.08$ &$550.33$ \\
                              & VCD                       &$148.58$ &$113.67$ &$133.00$ &$164.92$ &$560.17$ \\
                              & ICD                       &$144.75$ &$116.44$ &$\textbf{135.17}$ &$162.75$ &$559.11$    \\        
                              & Ours                      &$\textbf{157.75}$ &$\textbf{118.75}$ &$132.83$ &$\textbf{168.75}$ &$\textbf{578.08}$    \\ 
                               \midrule\bottomrule                             

\end{tabular}
\caption{Results on the hallucination subset of MME. 
The best performances within each setting are \textbf{bolded}.}
\label{tab:mme}
%\vspace{-0.2cm}
\end{table}
For the MME full set, Figure.~\ref{fig:mme_full} illustrates the performance of LLaVA 1.5 with normal decoding, VCD, and our method. Our method generally outperforms the other methods in 10 sub-tasks of 14. Notably, our method demonstrates a strong capability for perception, outperforming VCD and ICD by a large margin. However, our approach is less effective in numerical calculations and text translation tasks, as these primarily rely on the language decoder's reasoning capability rather than visual content comprehension. Nevertheless, our method maintains comparable performance in recognition tasks compared to existing approaches. Overall, these findings illustrate that our method enhances the general functionality of LVLMs while also effectively mitigating hallucinations.

\subsection{Ablation Study and Analysis}
In this section, we conduct ablations to demonstrate the effectiveness of our method. Without specification, our ablation study is done on the popular setups of POPE and the MS-COCO validation set for CHAIR.

\noindent \textbf{Impact of individual proposed modules.}
We conducted an ablation study to assess the effectiveness of each individual module, including IMD and CMPC, within our CMAC framework.
As illustrated in Table~\ref{tab:ablation_main}, IMD significantly improves the performance by 5.12\% on Accuracy and 5.34\% on F1 for VQA task, underscoring the significance of IMD on mitigating hallucination.
The combination of IMD and CMPC demonstrates a 1.02\% performance improvement of F1, which validates the effectiveness of CMPC in alleviating position bias. For image-captioning tasks, IMD boosts the performance on both $\text{CHAIR}_s$. Based on IMD, CMPC slightly enhances the performance, showcasing the efficacy of CMPC for long sequence generation tasks.

\begin{table}[h]
\centering
\fontsize{8.5}{8.5}\selectfont
\begin{tabular}{@{}cccccc@{}}
\toprule
\multirow{2}{*}{\textbf{IMD}}        & \multirow{2}{*}{\textbf{CMPC}} & \multicolumn{2}{c}{\textbf{POPE}}                                   & \multicolumn{2}{c}{\textbf{CHAIR}}            \\
                              &                           & \multicolumn{1}{c}{Accuracy} & \multicolumn{1}{c}{F1 Score} & \multicolumn{1}{c}{$\text{CHAIR}_i$} & \multicolumn{1}{c}{$\text{CHAIR}_s$}                  \\ \midrule
       &           & 79.98 & 79.34 & 55.6 & 17.8  \\
       \checkmark    &         &  85.10 &  84.68  &  48.2  & 13.6 \\
               &     \checkmark       & 82.31 & 81.86 & 54.2   &  16.2  \\     
   \checkmark    & \checkmark     & 86.04  & 85.70 & 47.0  &  12.7  \\       
                               \bottomrule                             

\end{tabular}
\caption{Ablation of the core components on POPE Popular and CHAIR for LLaVA 1.5.}
\label{tab:ablation_main}
%\vspace{-0.2cm}
\end{table}

\noindent \textbf{Ablation of distortion used in our method}.
In Table \ref{tab:ablation_distortion}, we show the performance of our method with different kinds of distortions. Specifically, 'Attention mask' indicates pruning significant attention weights by the conventional attention mask strategy. 'Value noise addition' indicates adding noise on value vectors similar to VCD~\cite{leng2024mitigating} instead of masking the value vectors to the mean value, which is denoted as 'Value mask'. Our method shows a better hallucination mitigation capability compared with attention masking, which demonstrates that altering the attention weights may lead to over- or under-estimation of hallucinations from image regions. In contrast, the value mask distortion in our method also surpasses the value noise addition, achieving a better generalization in different kinds of tasks.
\begin{table}[h]
\centering
\setlength\tabcolsep{3pt}
\fontsize{8.5}{8.5}\selectfont
\begin{tabular}{@{}ccccc@{}}
\toprule
\multirow{2}{*}{\textbf{Distortion}}        &   \multicolumn{2}{c}{\textbf{POPE}}                                   & \multicolumn{2}{c}{\textbf{CHAIR}}            \\
                                         & \multicolumn{1}{c}{Accuracy} & \multicolumn{1}{c}{F1 Score} & \multicolumn{1}{c}{$\text{CHAIR}_i$} & \multicolumn{1}{c}{$\text{CHAIR}_s$}                  \\ \midrule
 Attention mask    & 84.92  &  85.21 & 48.4   &  13.9  \\ 
Value noise addition     & 85.72  &  85.60 & 49.5 &  13.8  \\ 
 Value mask  & 86.04  & 85.70 & 47.0  &  12.7  \\       
                               \bottomrule                                   
\end{tabular}
\caption{Ablation of different kinds of distortion used in our method.}
\label{tab:ablation_distortion}
%\vspace{-0.2cm}
\end{table}

\begin{figure}[h]
  \centering
  \begin{subfigure}{0.22\textwidth}
    \centering
    \includegraphics[width=\textwidth, trim=0 0 0 0]{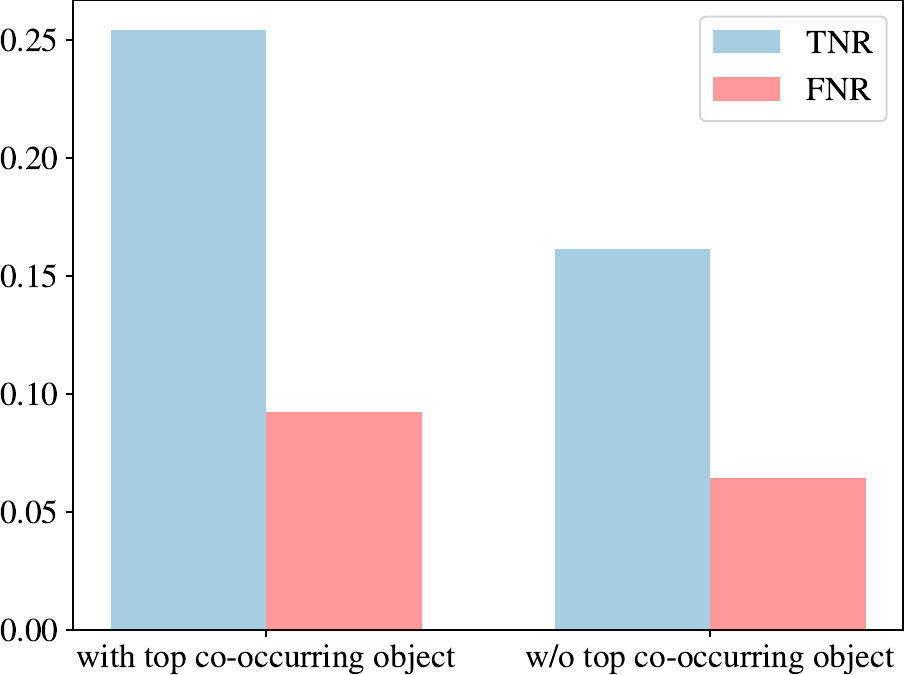}
    % \caption{Feature distributions generated by BARON.}
    \caption{}
    \label{fig:co-exist-a}
  \end{subfigure}
  \hfill
  \begin{subfigure}{0.22\textwidth}
    \centering
    \includegraphics[width=\textwidth, trim=0 0 0 0]{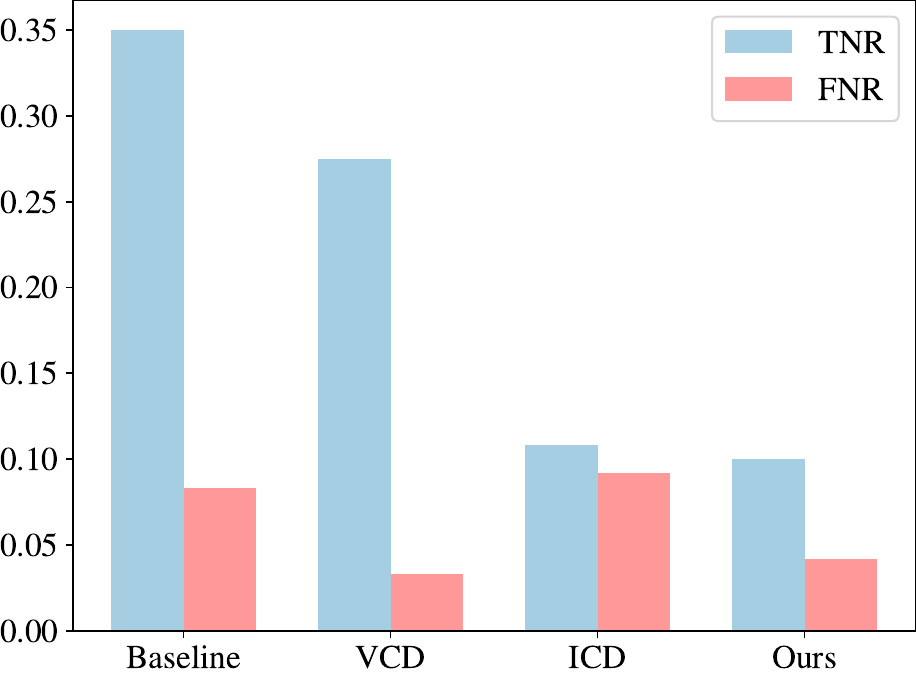}
    % \caption{Feature distributions generated by our LBP.}
    \caption{}
    \label{fig:co-exist-b}
  \end{subfigure}

% \begin{figure}[t]
% \centering
% % \vspace{-0.25cm}
% % \includegraphics[width=\textwidth, trim=0 0 0 5, clip]{images/pipeline_final.pdf}
% \includegraphics[width=0.99\linewidth]{imgs/bar_concat_v2.pdf}
% % \vspace{-3mm}
% \vspace{-0.25cm}

% \vspace{-0.25cm}
  \caption{The comparison of the hallucination rate of LLaVA 1.5 on the POPE dataset. 'TNR' and 'FNR' denote the true negative rate and the false negative rate of VQA, respectively. (a) The hallucination rate of the existence of objects with and without the co-existence with their top co-occurring object in the image. (b) The hallucination rate of the object's existence for different decoding methods with their top co-occurring object. Concretely, we estimate the mean hallucination rate on 5 pairs of objects with a high object existence rate.
}
  \label{fig:co-exist}
\end{figure}

\noindent \textbf{Analysis of hallucination with statistically significant cross-modal object existence.}
To assess the effectiveness of our method in mitigating hallucinations caused by spurious inter-modality correlations, we evaluate the hallucination rates of LLaVA 1.5 on the MS-COCO and A-OKVQA subsets of the POPE dataset in scenarios with statistically significant cross-modal object existence, which often induces spurious correlations. Figure~\ref{fig:co-exist}(a) illustrates that the statistically significant co-existence leads to a serious hallucination both on the true negative rate and the false negative rate. This finding indicates that spurious inter-modality correlations can cause both the neglect of relevant regions of interest and an over-reliance on irrelevant regions. Figure~\ref{fig:co-exist}(b) demonstrates that our method effectively reduces both true negative and false negative hallucinations stemming from co-existence, underscoring its ability to mitigate hallucinations  from spurious inter-modality correlations.

\noindent \textbf{Analysis of hallucination with position bias.} We investigate the distribution of hallucinations in LLaVA-1.5 with respect to token position. It can be observed from Table.\ref{tab:ablation_cmpc} that the hallucinations occur more frequently in the head portion of the image tokens, a phenomenon caused by the inherent limitations of the RoPE (Rotary Position Embedding) mechanism. By contrast, with the integration of our Cross-Modal Position Calibration (CMPC), the model not only exhibits a significant reduction in hallucinations but also achieves a more even distribution of hallucinations across different parts of image tokens. These findings highlight the effectiveness of our CMPC strategy in mitigating position bias and reducing hallucinations in multimodal models.

\begin{table}[h]
\centering
\fontsize{8.5}{8.5}\selectfont
\begin{tabular}{cccc}
\toprule
\textbf{Method} & \textbf{Head Part} & \textbf{Middle Part} &\textbf{ Tail Part} \\
\midrule
Baseline  & 28\% & 21\%& 15\% \\
Ours     &    11\% & 13\% & 10\%   \\   
                               \bottomrule                             

\end{tabular}
\caption{The effect of our method in mitigating the positional bias of the hallucination occurrence frequency.}
\label{tab:ablation_cmpc}
%\vspace{-0.2cm}
\end{table}
% \begin{figure}[h]
%     \centering
%     \includegraphics[width=0.8\linewidth, trim=0 0 0 0, clip]{imgs/cmpc_performance.pdf}
%     \caption{The hallucination distribution on the image token position before and after the application of Cross-Modal Position Calibration (CMPC). 
%     }
%     \label{fig:distribution_loc}
% \end{figure}
\section{Conclusion}
In this paper, we have introduced CMAC, a novel framework addressing the hallucination problem of LVLMs with contrastive decoding. In this approach, we propose the Inter-Modality Decoding which estimates a distorted distribution favored hallucination by selectively masking the value vectors associated with high cross-modal attention weights, alleviating both the spurious inter-modality correlation and the uni-modal over-reliance. Furthermore, a Cross-Modal Position Calibration module is introduced to alleviate the overlook of essential visual content caused by position bias.  Extensive experiments conducted on diverse benchmarks and LVLMs confirm the efficacy of the proposed method. 
% In this paper, we have introduced LBP, a novel framework addressing the challenges of open-vocabulary object detection. 
% In this approach, learning background prompts is proposed to harness explored implicit background knowledge. This can enhance the capacity of the detecter to recognize both base and novel categories during inference. 
% To achieve this, we have devised three essential modules: Background Category-specific Prompt, Background Object Discovery, and Inference Probability Rectification. These modules collectively empower the detector to discover, represent, and leverage implicit object knowledge explored from background proposals.

\subsection*{Acknowledgments}
This work is supported in part by the National Key
R\&D Program of China (NO.~2024YFB3908503 and 2024YFB3908500), and in part by the National Natural Science Foundation of China (NO.~62322608 and NO.~62495081).

{
    \small
    \bibliographystyle{ieeenat_fullname}
    \bibliography{main}
}

\clearpage
\appendix
% \clearpage
\setcounter{page}{1}
% \maketitlesupplementary

\section{More Comparisons with Existing Methods}
\noindent \textbf{Detailed Comparison on POPE under VCD Settings}.
We report the detailed performance on POPE in Table~\ref{tab:POPE} with nucleus sampling under VCD settings for further comparison. As presented in the table, CMAC leads to performance improvement on both precision and recall on most of setups, which demonstrates the generalization capability of CMAC to generate both precise and comprehensive responses.

\begin{table*}[t]
\centering
\small % Smaller font size
\setlength\tabcolsep{3pt} % Baseline value: 6pt
\fontsize{9}{9}\selectfont
\begin{tabular}{lllcccccccc}
\toprule
\multirow{2}{*}{\textbf{Dataset}} & \multirow{2}{*}{\textbf{Setup}} & \multirow{2}{*}{\textbf{Method}}& \multicolumn{4}{c}{\textbf{LLaVA 1.5}} & \multicolumn{4}{c}{\textbf{InstructBLIP}} \\
\cmidrule(lr){4-7} \cmidrule(lr){8-11}
 &  &  & {\textbf{Accuracy}} & {\textbf{Precision}} & {\textbf{Recall}} & {\textbf{F1 Score}} & {\textbf{Accuracy}} & {\textbf{Precision}} & {\textbf{Recall}} & {\textbf{F1 Score}} \\
\midrule
\multirow{12}{*}{MS-COCO} & \multirow{4}{*}{Random} & \textit{Baseline} & 83.29 & 92.13 & 72.80 & 81.33 & 80.71 & 81.67 & 79.19 & 80.41\\
 &  & \textit{VCD} & 87.73 & 91.42 & 83.28 & 87.16 & 84.53 & 88.55 & 79.32 & 83.68 \\
 &  & \textit{ICD} & 85.17 & \textbf{95.99} & 73.40 & 83.19 & \underline{86.43} & 92.01 & \textbf{80.73} & \underline{85.61} \\

&  & \textit{Ours} & \textbf{89.23} & 93.50 & \textbf{84.33} & \textbf{88.68} & \textbf{86.96} & \textbf{93.97} & 79.00 & \textbf{85.83} \\

  \cmidrule(lr){3-11} 
  & \multirow{4}{*}{Popular} & \textit{Baseline} & 81.88 & 88.93 & 72.80 & 80.06 &	78.22&	77.87	&78.85&	78.36\\
 &  & \textit{VCD} & 85.38  & 86.92 &	\textbf{83.28}	& 85.06	&81.47&	82.89&	79.32&	81.07 \\
 &  & \textit{ICD} &82.33 & 82.25 & \underline{82.47} & 82.36  &	82.93 &	84.45 &	\textbf{80.73}	& \underline{82.55} \\

&  & \textit{Ours} & \textbf{86.90} & \textbf{92.28} & 80.53 & \textbf{86.01} & \textbf{84.13} & \textbf{88.38} & 78.60 & \textbf{83.20} \\
  \cmidrule(lr){3-11} 
  & \multirow{4}{*}{Adversarial} & \textit{Baseline} & 78.96 & 83.06 & 72.75 & 77.57	&75.84&	74.30	&79.03&	76.59 \\
 &  & \textit{VCD} & 80.88 & 79.45	& \textbf{83.29}	& 81.33	&79.56	&79.67	&79.39	&79.52 \\
 &  & \textit{ICD} & 81.17 & \textbf{87.13} & 73.13 & 79.52	&80.87	&80.95	& \textbf{80.73} & \underline{80.84} \\

 &  & \textit{Ours} & \textbf{83.23} & 85.28 & 80.33 & \textbf{82.73} & \textbf{82.57} & \textbf{86.10} & 77.67 & \textbf{81.67} \\
 \midrule
 
 \multirow{12}{*}{A-OKVQA} & \multirow{5}{*}{Random} & \textit{Baseline} & 83.45 & 87.24 & 78.36 & 82.56 &	80.91&	77.97	&86.16&	81.86 \\
 &  & \textit{VCD} & 86.15 & 85.18	& \textbf{87.53}	& 86.34	&84.11	&82.21&	87.05&	84.56 \\
 &  & \textit{ICD} & 86.20 & \textbf{91.07} & 80.27 & 85.33	&\underline{85.82}	&83.80	&\textbf{88.94}& \underline{86.29} \\

&  & \textit{Ours} & \textbf{88.53} & \underline{90.19} & 86.47 & \textbf{88.29} & \textbf{87.03} & \textbf{88.71} & 84.87 & \textbf{86.75} \\
  \cmidrule(lr){3-11} 
  & \multirow{4}{*}{Popular} & \textit{Baseline} & 79.90	& 80.85 &	78.36&	79.59&	76.19&	72.16&	85.28&	78.17 \\
 &  & \textit{VCD} & 81.85 & 78.60 &	\textbf{87.53} &	82.82	&79.78	&76.00	&87.05	&81.15 \\
 &  & \textit{ICD} & 82.63 & 84.25 & 80.27 & 82.21	&\underline{81.64}	&78.50	&\textbf{88.77}& \textbf{83.32}\\
% &  & \textit{+AvisC} & \textbf{85.03} & \textbf{90.08} & 78.73 & 84.03  & \textbf{82.47} & \textbf{81.79} & 83.53 & 82.65 \\
&  & \textit{Ours} & \textbf{85.73} & \textbf{86.66} & 84.47 & 85.55 & \textbf{82.60} & \textbf{80.64} & 85.80 & \underline{83.14} \\
  \cmidrule(lr){3-11}

  & \multirow{4}{*}{Adversarial} & \textit{Baseline} & 74.04&	72.08&	78.49&	75.15&	70.71&	65.91&	85.83&	75.56 \\
 &  & \textit{VCD} & 74.97 &	70.01&	\textbf{87.36} &	77.73&	\underline{74.33} &	 \underline{69.46} &	86.87&	77.19 \\
 &  & \textit{ICD} & 77.20 & 75.47 & 80.60 & 77.95	&\underline{74.42}	& \underline{70.24}	&\textbf{88.93}	&\textbf{78.48} \\
 % &  & \textit{+AvisC} & \textbf{79.27} & \textbf{79.58} & 78.73 & 79.16  & \textbf{76.47} & \textbf{73.16} & 83.60 & 78.03 \\
&  & \textit{Ours} & \textbf{79.13} & \textbf{76.45} & 84.20 & \textbf{80.14} & \textbf{75.70} & \textbf{71.38} & 85.80 & \underline{77.93} \\
 \midrule
 \multirow{12}{*}{GQA} & \multirow{4}{*}{Random} & \textit{Baseline} & 83.73	&87.16	&79.12	& 82.95&	79.65&	77.14	&84.29&	80.56 \\
 &  & \textit{VCD} & 86.65	&84.85	&\underline{89.24}	& 86.99 &	83.69&	81.84&	\textbf{86.61}	&84.16\\
 &  & \textit{ICD} & 85.73 & \textbf{90.79} & 79.53 & 84.79 &\underline{85.10}	&84.21&	\underline{86.40}	& \underline{85.29} \\

&  & \textit{Ours} & \textbf{89.53} & \underline{90.62} & \textbf{88.20} & \textbf{88.80}  & \textbf{86.03} & \textbf{87.67} & 83.87 & \textbf{85.72} \\

  \cmidrule(lr){3-11} 
  & \multirow{4}{*}{Popular} & \textit{Baseline} & 78.17& 	77.64& 	79.12& 	78.37& 	73.87& 	69.63& 	84.69& 	76.42 \\
 &  & \textit{VCD} & 80.73	& 76.26 & 	\textbf{89.24} & 82.24& 	78.57 & 	74.62& 	\underline{86.61} & 	80.17 \\
 &  & \textit{ICD} & 79.87 & 80.07 & 79.53 & 79.80& 	\underline{78.80}	& 75.15  & \textbf{87.53}	& \textbf{80.87} \\

&  & \textit{Ours} & \textbf{85.50} & \textbf{85.24} & 85.87 & \textbf{85.55}  &\textbf{79.87} & \textbf{77.28} & 84.60 & \underline{80.78} \\
  \cmidrule(lr){3-11} 
  & \multirow{4}{*}{Adversarial} & \textit{Baseline} & 75.08& 	73.19& 	79.16	& 76.06	& 70.56	& 66.12	& 84.33	& 74.12 \\
 &  & \textit{VCD} & 76.09	& 70.83	& \textbf{88.75}	& 78.78 & 	75.08 & 	70.59	& 85.99 & \underline{77.53} \\
 &  & \textit{ICD} & 77.60 & 76.04 & 80.60 & 78.25 & 	75.17	& 70.59& 	\textbf{86.27}& 	\underline{77.65} \\

&  & \textit{Ours} & \textbf{81.87} & 79.18 & 86.47 & \textbf{82.66}  & \textbf{76.83} & \textbf{73.94} & 82.87 & \textbf{78.15} \\
\bottomrule
\end{tabular}
\caption{\textbf{Results on discrimination hallucination benchmark POPE.} The Baseline method denotes the standard decoding. The best performances within each setting are \textbf{bolded}. Comparable ($\pm 1.0$) but not the best performances are \underline{underlined}. We implemented ICD for LLaVA 1.5 in our evaluation setup.
}
\label{tab:POPE}
\end{table*}

\noindent \textbf{GPT-4o Assisted Evaluation}.
To assess the effectiveness of our method in mitigating hallucination beyond object existence during long-sequence generation, we conducted open evaluations of LVLM-generated responses using GPT-4o, as presented in Table~\ref{tab:gpt4o}. Following PAI~\cite{liu2024paying}, we sampled 50 images from the COCO dataset for image captioning tasks. Unlike PAI, which employs GPT4-V, we utilized the recently released GPT-4o to evaluate generated responses in terms of accuracy and detail. The GPT-4o prompts were designed in alignment with the structure in PAI~\cite{liu2024paying}. We compare our method with the baseline method and existing decoding methods, including VCD~\cite{leng2024mitigating} and ICD~\cite{wang2024mitigating}. Our proposed method achieved superior GPT-4o scores compared to baseline and existing decoding approaches,  reflecting its robustness under a more comprehensive evaluation framework. These results highlight the enhanced capability of our approach to generate more accurate and detailed responses for image captioning tasks.

\begin{table}[h]
\centering
\begin{tabular}{@{}ccc@{}}
\toprule
\textbf{Method} &  \textbf{Accuracy} $\uparrow$ &  \textbf{Detailedness} $\uparrow$ \\ \midrule
  Baseline     &      5.38     &  5.88  \\
  Ours     &       6.65    &  6.45 \\ \midrule
    VCD     &      5.83     & 5.93 \\
  Ours     &        6.57   &  6.54\\ \midrule
   ICD    &        6.03   & 6.48  \\    
    Ours   &      6.61    & 6.52 \\   
 \bottomrule                             

\end{tabular}
\caption{ The results of LLaVA 1.5 for GPT-4o assisted evaluation. All the metrics are on a scale of 1 to 10. }
\label{tab:gpt4o}

\end{table}

\noindent \textbf{Evaluation with  LLaVA-OneVision}.
To further assess the generalization capability of our method, we apply our method to the latest LLaVA-OneVision(llava-onevision-qwen2-7b-ov) on the POPE dataset. The results are reported in Table~\ref{tab:llava_onevision}. It shows that our method outperforms VCD and consistently enhances the performance of  LLaVA-OneVision, underscoring the generalization of our method.

\begin{table*}[t] 
\centering
\footnotesize
\begin{tabular}{ll|cc|cc|cc||cc|cc|cc}
\bottomrule
% --- Header Row 1: Decoding Strategy ---
& & \multicolumn{6}{c}{\textbf{Sampling}} & \multicolumn{6}{c}{\textbf{Greedy}} 

\\
\cmidrule(lr){3-8} \cmidrule(lr){9-14}

% --- Header Row 2: Setting ---
& & \multicolumn{2}{c|}{\textbf{Random}} & \multicolumn{2}{c|}{\textbf{Popular}} & \multicolumn{2}{c||}{\textbf{Adversarial}} & \multicolumn{2}{c|}{\textbf{Random}} & \multicolumn{2}{c|}{\textbf{Popular}} & \multicolumn{2}{c}{\textbf{Adversarial}} \\
\cmidrule(lr){3-4} \cmidrule(lr){5-6} \cmidrule(lr){7-8} \cmidrule(lr){9-10} \cmidrule(lr){11-12} \cmidrule(lr){13-14}

% --- Header Row 3: Metrics ---
\textbf{} & \textbf{Method} & Acc & F1  & Acc & F1  & Acc & F1 & Acc & F1  & Acc & F1  & Acc & F1 \\ 
\hline \hline

% --- QWEN-VL Data ---
\multirow{3}{*}{\textbf{LLaVA-OV}}
& Baseline & 87.18 & 86.07 & 84.07 & 83.40 & 80.99 & 80.90 & 90.10 & 89.30 & 87.81 & 87.17 & 84.17 & 84.02 \\
% & ICD     & 86.45 & 85.80 & 83.41 & 83.27 & 81.04 & 81.33 & 86.49 & 85.86 & 83.80 & 83.71 & 80.96 & 81.27 \\
& VCD      & 88.69 & 87.78 & 86.18 & 85.56 & 82.89& 82.38 & 90.28 & 89.44 & 88.57 & 87.83 & 84.20 & 83.86 \\
& \textbf{Ours}  & \textbf{89.52} & \textbf{88.83} & \textbf{87.32} & \textbf{86.71} & \textbf{83.69} & \textbf{83.40} & \textbf{90.45} & \textbf{89.50} & \textbf{89.12} & \textbf{88.38} & \textbf{84.93} & \textbf{84.67} \\
\bottomrule
\end{tabular}
\caption{\textbf{Results on discrimination hallucination benchmark POPE.} The Baseline method denotes the standard decoding. The best performances within each setting are \textbf{bolded}. "Acc" and "F1" denote the Accuracy and F1 scores, respectively. }
\label{tab:llava_onevision}
\end{table*}

\noindent \textbf{Evaluation on MMBench}.
We also test the results on the general ability benchmark MMBench on LLaVA-1.5. As in Table~\ref{tab:mmbench}, our method achieves a clear improvement on the general ability of LVLMs.

\begin{table}[h]
\centering

\begin{tabular}{@{}cc@{}}
\toprule
\textbf{Method}       &   \textbf{MMBench Score}                                            \\ \midrule
Baseline   &  63.9\\ 
Ours & 64.7 \\ 
                               \bottomrule                             
\end{tabular}
\caption{Results of LLaVA1.5 on general ability benchmark MMbench.}
\label{tab:mmbench}

\end{table}

\noindent \textbf{Detailed Comparison on CHAIR}.
The $\textit{max new token}$ parameter plays a critical role in the evaluation of the CHAIR metric by restricting the maximum length of generated responses. In the main text, we present results under the settings $\textit{max new token} = 1024$ and $\textit{top p} = 1$. Additionally, following \cite{huang2024opera, woo2024ritual}, we also report results for $\textit{max new token} = 64$ in Table~\ref{tab:chair_more}. As illustrated in the table, the $\textit{max new token}$ value significantly influences the performance of LVLMs on the CHAIR metric. Nevertheless, our method outperforms other methods by a clear margin. It shows our method consistently demonstrates superior hallucination mitigation capabilities, further validating its robustness and effectiveness under a strict constraint on the length of generated responses.
\begin{table}[]
\centering
\small
\begin{tabular}{ccccc}
\toprule
\textbf{Method} &   $\textbf{CHAIR}_i \downarrow$    &    $\textbf{CHAIR}_s\downarrow$   &  \textbf{Recall} $\uparrow$    &     \textbf{F1 Score} $\uparrow$  \\ \midrule
 Baseline  &    25.4   &   9.1    &   56.7    &   69.8      \\
 VCD &    22.0   &   6.9   &   62.0    &     74.4    \\
 ICD &    22.2  &    7.8  &   61.1   &   73.5      \\
 Ours  &   20.3   &   5.9   &   61.9    &  74.7       \\ \bottomrule
\end{tabular}
\caption{Results of CHAIR on the MS-COCO validation set with the $\textit{max new token} = 64$.}
\label{tab:chair_more}
\end{table}

% \noindent \textbf{Comparison with Beam Search.}
% We also compare our method on LLaVA 1.5 with beam search on CHAIR, which is presented in Table.~\ref{}.

% \input{tables/GPT4O_prompt}

\noindent \textbf{Inference Time Analysis}.
\begin{table}[h]
\centering
% \setlength\tabcolsep{3pt}
% \fontsize{8.5}{8.5}\selectfont
\begin{tabular}{@{}cc@{}}
\toprule
\textbf{Method}       &   \textbf{Inference speed (tokens/s)}                                            \\ \midrule
Baseline   &  12.93 \\ 
VCD     & 7.04  \\ 
ICD &  6.90 \\      
Ours & 7.75 \\ 
                               \bottomrule                             

\end{tabular}
\caption{Ablation of inference speed in our method.}
\label{tab:time}
%\vspace{-0.2cm}
\end{table}

To validate the effectiveness of our method on the inference speed, we estimate the inference time consumption of MLLMs employing different decoding strategies in the POPE dataset in Table~\ref{tab:time}. The results demonstrate that our proposed method achieves faster inference speeds compared to existing methods such as VCD and ICD. This improvement stems from CMAC's ability to directly derive the attention weights and key-value vectors for the distorted forward process from the original forward process, avoiding redundant computations. 

\section{More Ablation and  Analysis}
\noindent \textbf{Analysis of ROPE with CMPC}.
To illustrate the advantages of our proposed CMPC, we first analyze the mechanism and limitations of Rotary Position Embedding (RoPE). RoPE encodes positional information by rotating query and key vectors according to their absolute positions. As shown in Equation~\ref{eq:rope2}, the rotation for a token at position $j$ and another at position $i$ is applied independently.
\begin{align}
\mathbf{R}_k^i =  &
\begin{pmatrix}
% \setstacktabbedgap{2pt}
\cos{(i\theta_k)}& -\sin{(i\theta_k)}\\
\sin{(i\theta_k})&\cos{(i\theta_k)} 
\end{pmatrix},  \quad  
\theta_k = \frac{1}{b^{2k/d}}, \nonumber \\
\mathbf{a}^{l}_{ji} \propto & (\mathbf{q}^j \mathbf{R}^j) (\mathbf k^i \mathbf{R}^i)^T = \mathbf{q}^j
\mathbf{R}^j (\mathbf{R}^i)^T (\mathbf k^i)^T  \nonumber\\ \quad & \quad\quad\quad\quad\quad\quad\quad  =  
\mathbf{q}^j
 (\mathbf{R}^{j-i})^T (\mathbf k^i)^T ,
\label{eq:rope2}
\end{align}
A crucial property of RoPE, derived in Equation~\ref{eq:rope2}, is that the final attention score only depends on the relative distance $(j-i)$. However, this elegant design faces a critical failure mode when processing sequences longer than the model's training context. The rotation angle $(j-i)\theta_k$, is proportional to this distance. In high-frequency dimensions (i.e., for small 
$k$, where $\theta_k$ is large), \textit{a large relative distance can cause the angle to exceed 
$2\pi$.}

\begin{figure}[h]
  \centering
  \setlength{\abovecaptionskip}{0.1cm}
  \begin{subfigure}{0.23\textwidth}
    \centering
    \includegraphics[width=\textwidth, trim=0 0 0 0]{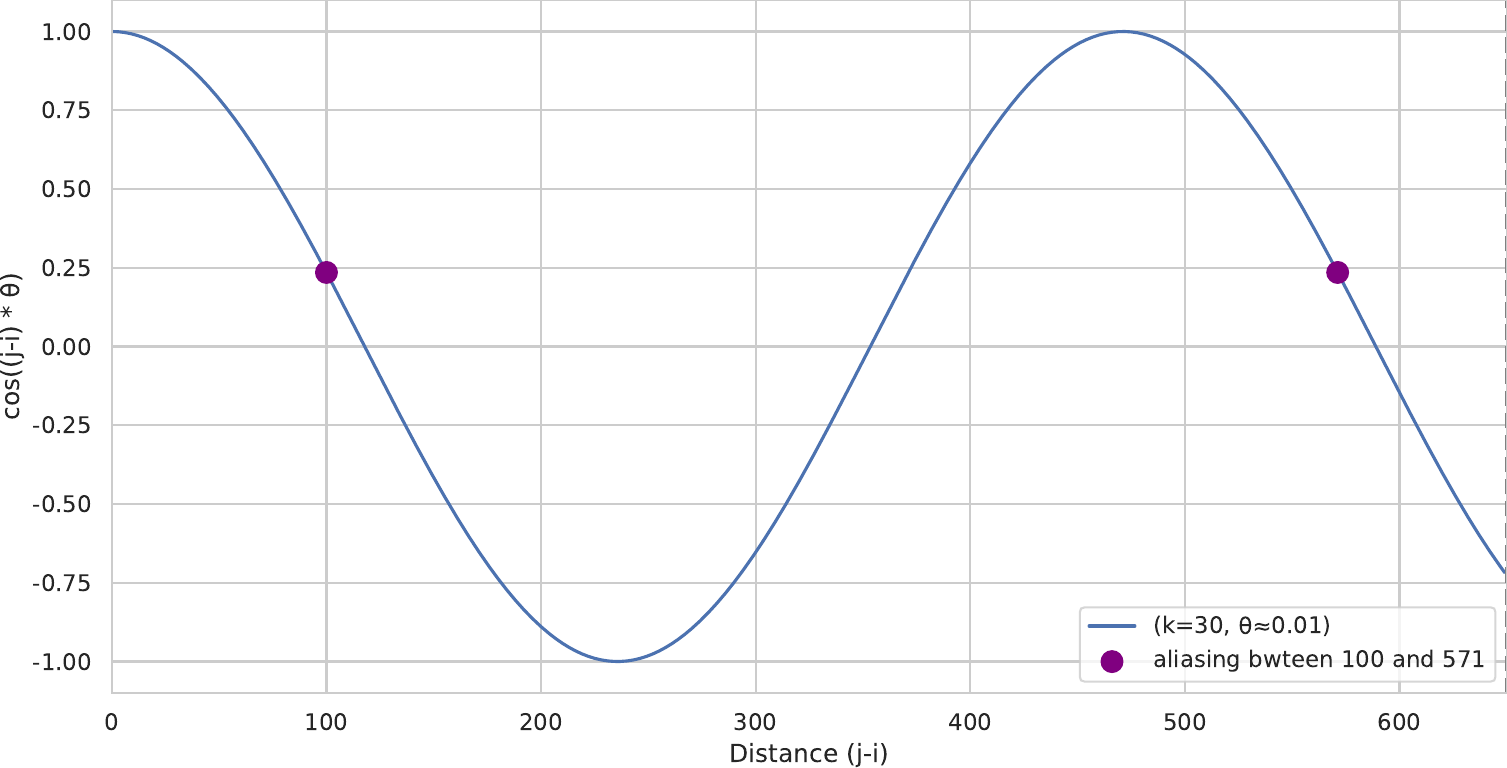}
    % \caption{Feature distributions generated by BARON.}
    \caption{without CMPC}
    \label{fig:alias-a}
  \end{subfigure}
  \hfill
  \begin{subfigure}{0.23\textwidth}
    \centering
    \includegraphics[width=\textwidth, trim=0 0 0 0]{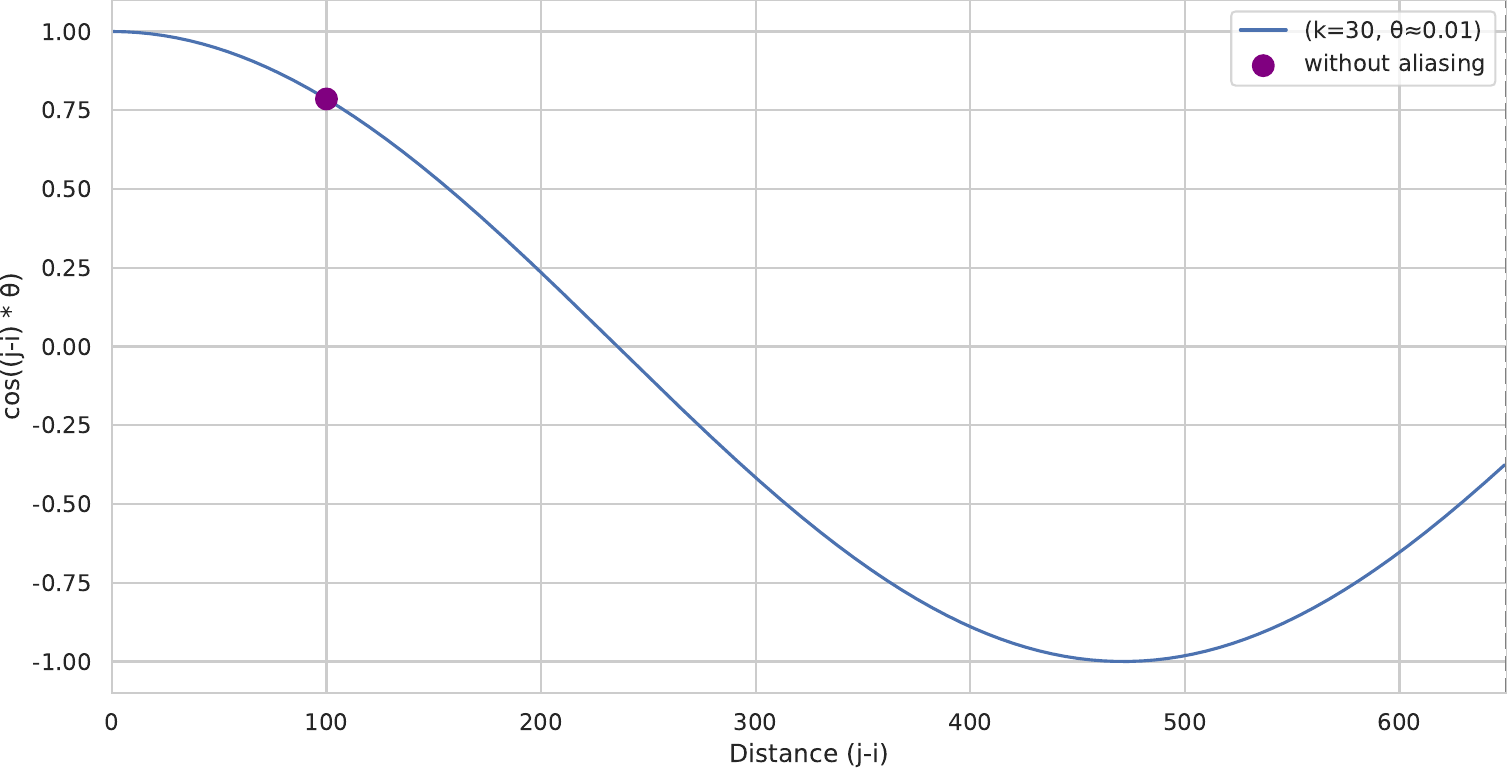}
    % \caption{Feature distributions generated by our LBP.}
    \caption{with CMPC}
    \label{fig:alias-b}
  \end{subfigure}

% \vspace{-0.25cm}
  \caption{The rotation value of ROPE under different token distances without (a) and with (b) the CMPC.
}
  \label{fig:alias}
\vspace{-0.5cm}
\end{figure}

\textit{Once the angle surpasses this $\pi$ threshold, the trigonometric functions are no longer monotonic, leading to ambiguity.} For instance, the model cannot distinguish a relative distance from another distance if their corresponding rotation angles map to the same cosine value, as shown in Figure~\ref{fig:alias}. This phenomenon, known as position aliasing, causes the model to assign erroneous attention weights to distant tokens, resulting in a catastrophic degradation of performance.

Our proposed CMPC alleviates this issue. By systematically scaling the position indices of all tokens, CMPC effectively compresses the large rotation angles to low angles that the model can correctly interpret. This enhances the model's representation capability by preventing position aliasing and preserving its ability to accurately model long-range dependencies.

\begin{table*}[ht]
\centering
\setlength\tabcolsep{3pt}
\fontsize{8}{8.5}\selectfont
\begin{tabular}{@{}lccccccc@{}}
\toprule
\multirow{2}{*}{\textbf{Method}}        &   \multicolumn{2}{c}{\textbf{POPE}}                                   & \multicolumn{5}{c}{\textbf{CHAIR}}            \\
                                         & \multicolumn{1}{c}{Accuracy} & \multicolumn{1}{c}{F1 Score} & \multicolumn{1}{c}{$\text{CHAIR}_i$} & \multicolumn{1}{c}{$\text{CHAIR}_s$} &  Recall $\uparrow$    &     F1 $\uparrow$  &  Len                 \\ \midrule
   Without CMPC & 85.10 &  84.68  &  48.2  & 13.6  & 75.8 & 80.7 & 101.0 \\ 
 GPC  & 86.57  & 87.12  & 61.6 & 17.1 & 81.8 & 82.3 & 146.0\\
 GPC + EOS enhancement  & 86.57 & 87.12& 44.8	& 13.8	& 75.7	& 80.6 &	102.3  \\
 CMPC  & 86.04  & 86.70 & 47.0  &  12.7 & 75.6&	81.0&	102.4 \\       
                               \bottomrule                             

\end{tabular}
\caption{Ablation of different variants of CMPC module in our method. 'GPC' denotes global position calibration, which directly scales up all the position indexes on all the attention weights. 'EOS enhancement' denotes directly enhancing the probability of the EOS tokens.}
\label{tab:ablation_cdar}
%\vspace{-0.2cm}
\end{table*}

\begin{figure}[h]
  \centering
  \setlength{\abovecaptionskip}{0.1cm}
  \begin{subfigure}{0.23\textwidth}
    \centering
    \includegraphics[width=\textwidth, trim=0 0 0 0]{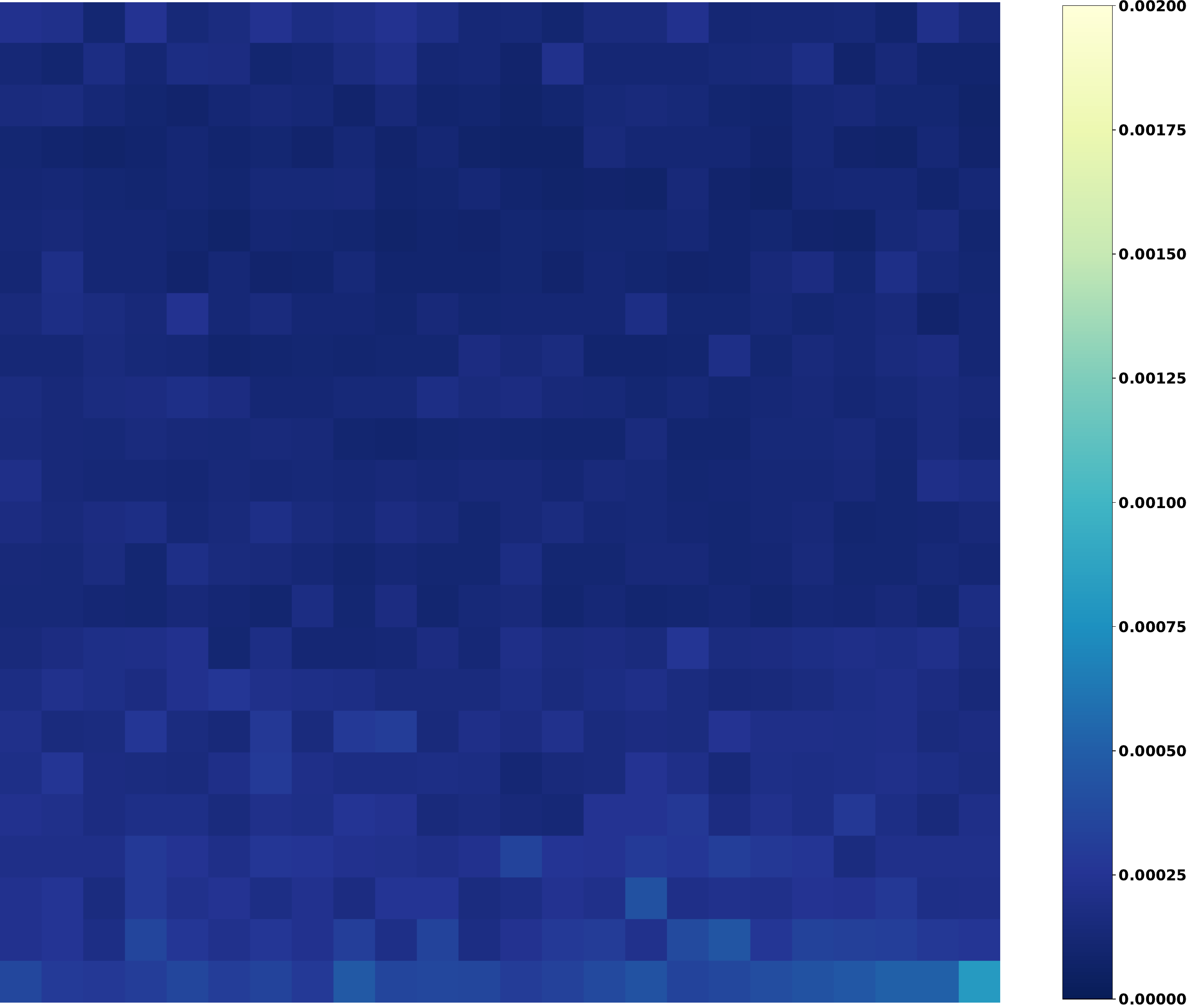}
    % \caption{Feature distributions generated by BARON.}
    \caption{without CMPC}
    \label{fig:attn-dis-a}
  \end{subfigure}
  \hfill
  \begin{subfigure}{0.23\textwidth}
    \centering
    \includegraphics[width=\textwidth, trim=0 0 0 0]{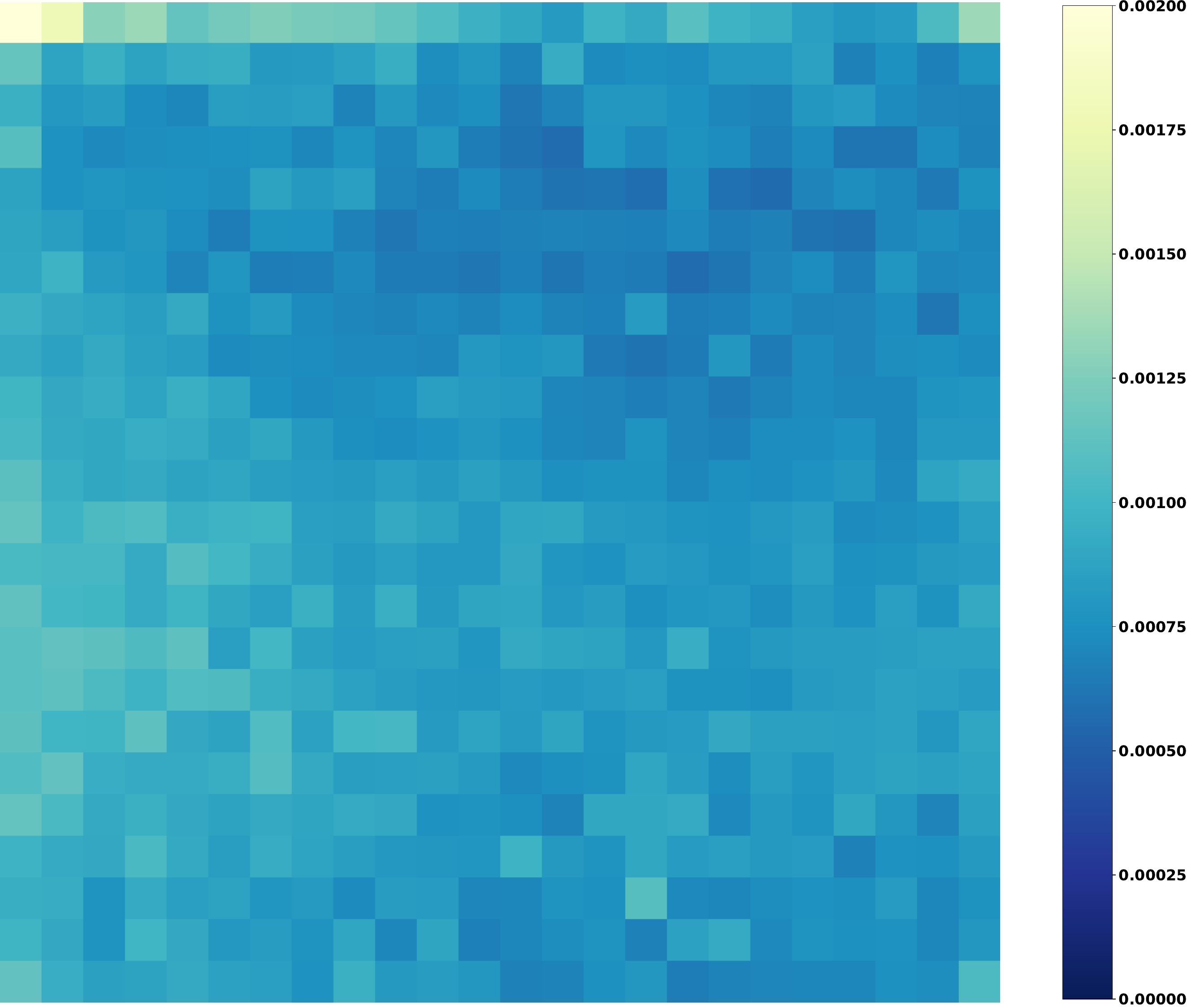}
    % \caption{Feature distributions generated by our LBP.}
    \caption{with CMPC}
    \label{fig:attn-dis-b}
  \end{subfigure}

% \vspace{-0.25cm}
  \caption{The attention distribution of text tokens to image tokens without (a) and  with (b) the position embeddings.
}
  \label{fig:attn-dis}
\vspace{-0.3cm}
\end{figure}

\begin{figure*}[h]
  \centering
\setlength{\abovecaptionskip}{0.1cm}
  \begin{subfigure}{0.47\textwidth}
    \centering
    \includegraphics[width=\textwidth, trim=0 0 0 0]{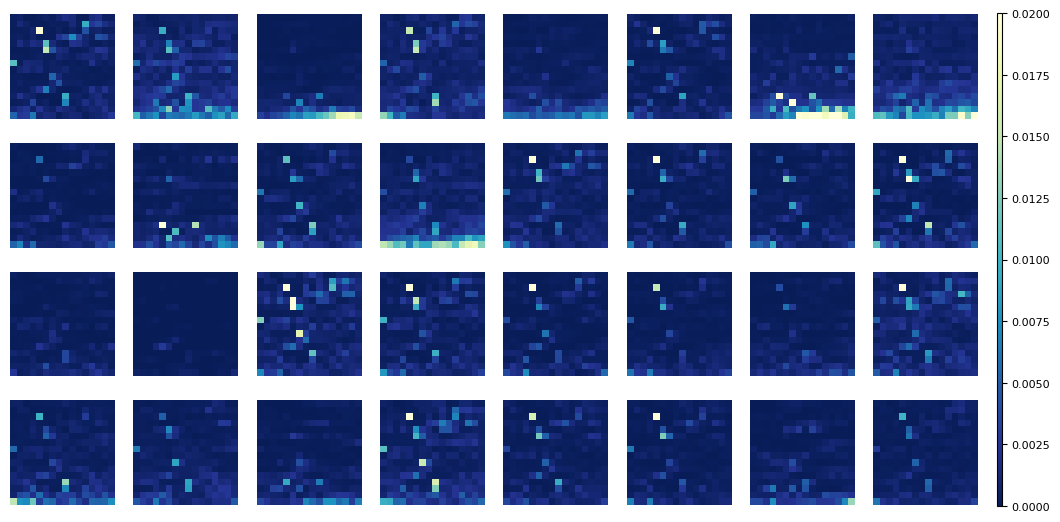}
    % \caption{Feature distributions generated by BARON.}
    \caption{without CMPC}
    \label{fig:attn-dis2-a}
  \end{subfigure}
  \hfill
  \begin{subfigure}{0.47\linewidth}
    \centering
    \includegraphics[width=\textwidth, trim=0 0 0 0]{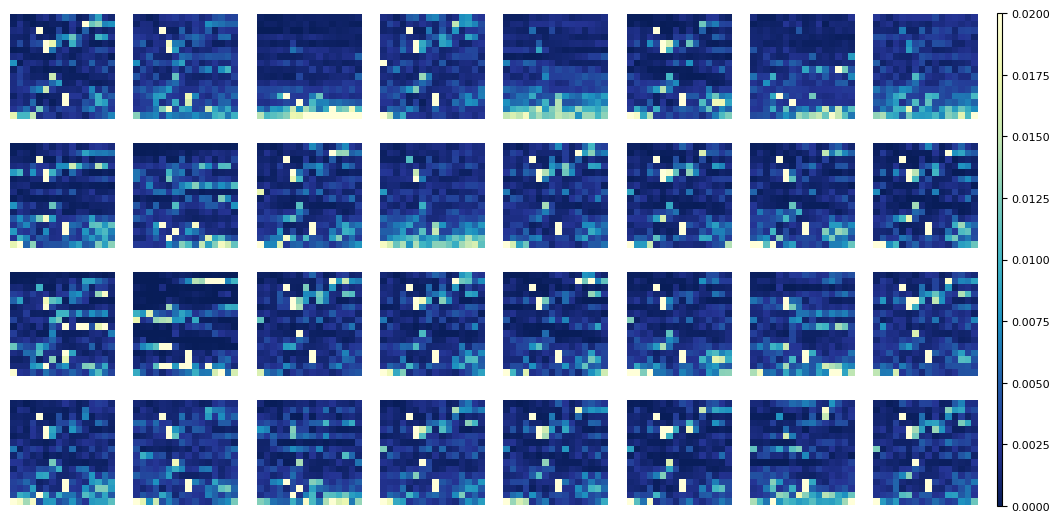}
    % \caption{Feature distributions generated by our LBP.}
    \caption{with CMPC}
    \label{fig:attn-dis2-b}
  \end{subfigure}

% \vspace{-0.25cm}
  \caption{The attention of different heads of text tokens to image tokens with (a) and without (b) the position embeddings.
}
  \label{fig:attn-dis2}
\vspace{-0.5cm}
\end{figure*}

\noindent \textbf{Analysis of ROPE with CMPC}.
 To further validate the effectiveness of CMPC, we analyze the attention distribution from text tokens to image tokens with and without position embeddings by visualizing the attention maps.   As depicted in Figure~\ref{fig:attn-dis}(a), when position embeddings are included, the text tokens predominantly focus on the latter portion of the visual content, neglecting other relevant regions. It shows that the position embeddings lead to position bias in the attention mechanism. Conversely, as in Figure~\ref{fig:attn-dis}(b), CMPC significantly enhances attention weights to image tokens.  We also visualize the attention maps of each head in Figure~\ref{fig:attn-dis2}. It can also be observed that the CMPC helps to encourage LVLMs to focus on the crucial information in images. These demonstrate that the CMPC effectively eliminates the position relation among different image tokens on the cross-modal part to mitigate the overlooking of critical visual information.
 
% We further analyze the attention distribution from text tokens to image tokens with and without position embeddings by visualizing the attention maps. Specifically, we estimate the average attention from the final text token to each image token using data from the POPE dataset. As depicted in Figure~\ref{fig:attn-dis}(a), when position embeddings are included, the text tokens predominantly focus on the latter portion of the visual content, neglecting other relevant regions. Conversely, as in Figure~\ref{fig:attn-dis}(b), removing the position embeddings significantly enhances attention to all image tokens. It shows that the position embeddings lead to position bias in the attention mechanism. CMPC eliminates the position relation among different image tokens on the cross-modal part to mitigate the overlooking of critical visual information.

\begin{figure*}[ht]
\centering
\includegraphics[width=0.95\textwidth]{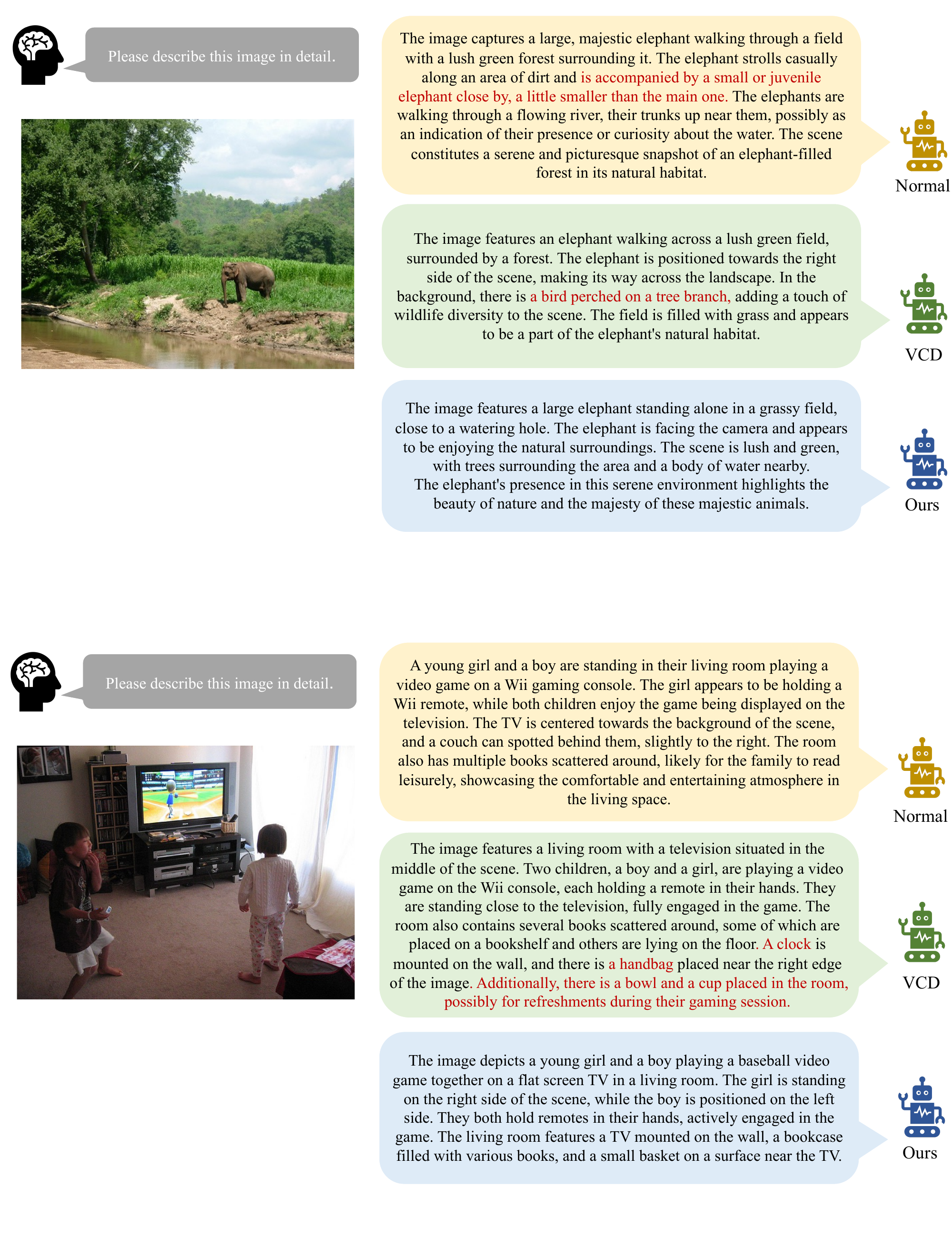}
\caption{Qualitative results of different methods on MS-COCO image captioning task. The hallucinated content is highlighted in red.}
\label{fig:qualitative_1}
\end{figure*}

% \begin{figure*}[t]
% \centering
% \includegraphics[width=0.95\textwidth]{imgs/qualitative_results_2.pdf}
% \caption{Qualitative results of different methods on MS-COCO image captioning task. The hallucinated content is highlighted in red.}
% \label{fig:qualitative_2}
% \end{figure*}
\begin{figure*}[ht]
\centering
\includegraphics[width=0.95\textwidth]{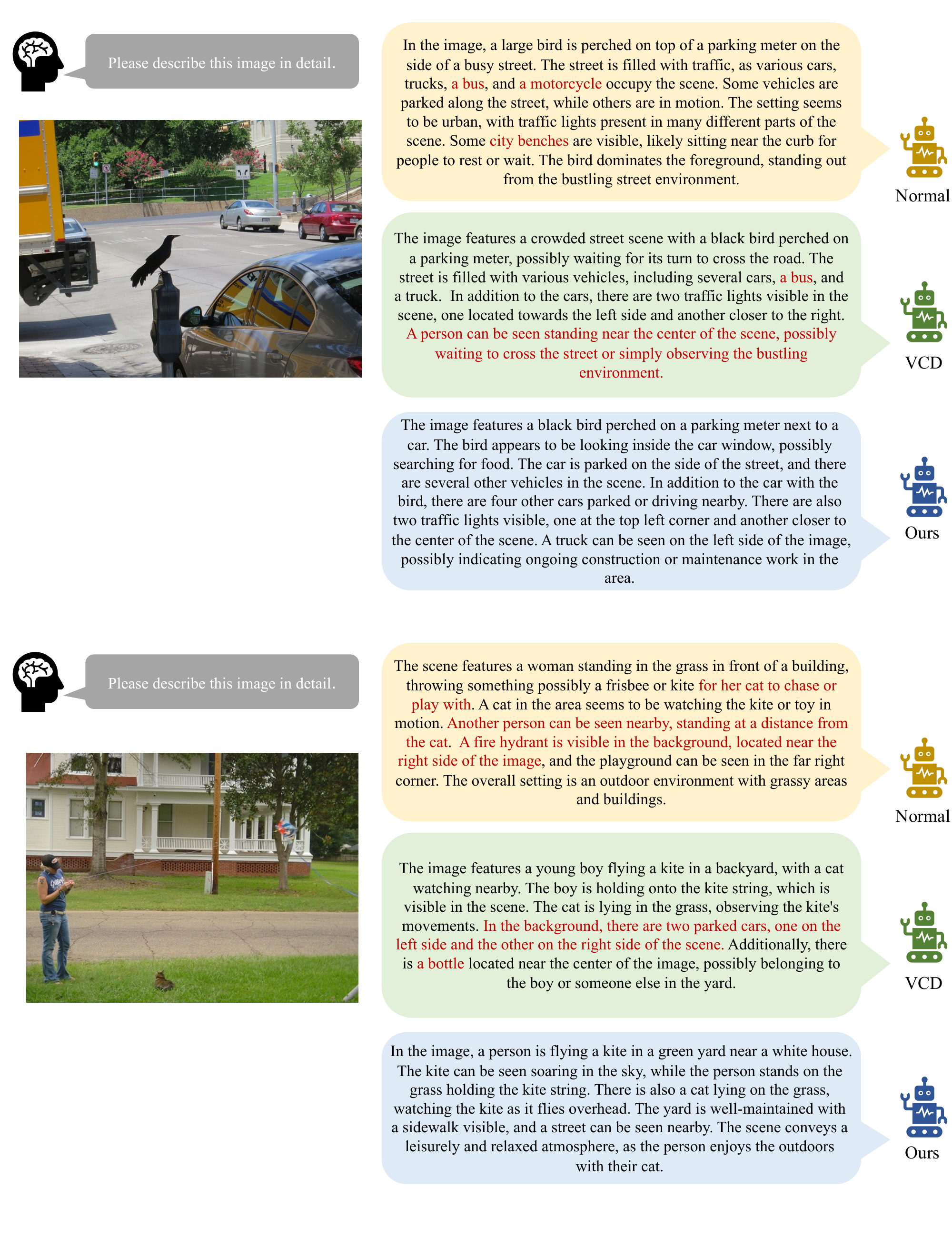}
\caption{Qualitative results of different methods on MS-COCO image captioning task. The hallucinated content is highlighted in red.}
\label{fig:qualitative_3}
\end{figure*}

\noindent \textbf{Ablation of variants of CMPC}. We conduct more ablation about the variants of the position embedding strategy in CMPC in Table~\ref{tab:ablation_cdar}. Note that "GPC" demotes global position calibration, which directly scales up all the position indexes on all the attention weights. We find that GPC performs even better than our method on the POPE dataset, indicating the position bias in uni-modal knowledge exchange. However, GPC scales the position indexes for text tokens, which makes the LVLMs hard to distinguish the length of generated text and induces more hallucinations. This issue can be alleviated by directly enhancing the probability of EOS tokens by a fixed value to normalize the length of the generated output. GPC + EOS enhancement leads to significant improvement on $\text{CHAIR}_i$ but slightly reduces the $\text{CHAIR}_s$. We consider it a variant of our method. Additionally, the refined position embeddings used in CMPC further lead to an improvement on both $\text{CHAIR}_i$ and $\text{CHAIR}_s$, which shows the significance of the global positional relation of the image content with other language tokens.

\section{Choices of Hyper-parameters}
In this section, we analyze the choices of hyperparameters used in the proposed approach, including $\alpha$, and $\gamma$. 

\noindent\textbf{Choice of $\alpha$}. $\alpha$ is the most important parameter for the contrastive decoding, which moderates the contrastive effect. To determine the value of $\alpha$, we compared model performance under different $\alpha$ for our method in Table~\ref {tab:ablation_alpha}. Compared with the performance under $\alpha = 1$, our method shows a better performance under a more aggressive setting $\alpha = 3$ and $\alpha = 5$ with nucleus sampling. However, $\alpha = 5$  leads to a performance degradation in greedy search (85.85 vs. 86.33 at $\alpha = 3$). Consequently, we selected $\alpha = 3$ as the default setting to ensure robust performance across different decoding strategies.

\begin{table}[ht]
\centering
\setlength\tabcolsep{3pt}
% \fontsize{8.5}{8.5}\selectfont
\small
\begin{tabular}{ccccc}
\toprule
\multirow{2}{*}{$\alpha$}         & \multicolumn{2}{c}{\textbf{Sampling}}                                   & \multicolumn{2}{c}{\textbf{Greedy search}}            \\
                             & \multicolumn{1}{c}{Accuracy $\uparrow$} & \multicolumn{1}{c}{F1 Score$\uparrow$} & \multicolumn{1}{c}{Accuracy$\uparrow$} & \multicolumn{1}{c}{F1 Score$\uparrow$}                  \\ \midrule
1      &  85.50 &  85.12  &  86.27  & 85.89 \\
3   & 86.04  & 85.70 & 86.33 &  86.00  \\     
5      & 86.42 & 86.33   &  85.85 & 85.52  \\       
                               \bottomrule                             
\end{tabular}
\caption{The performance of our method  with the different values of $\alpha$ on LLaVA 1.5 under POPE popular.}
\label{tab:ablation_alpha}
\end{table}

\noindent\textbf{Choice of $\gamma$}.
To better select $\gamma$, we present its performance under different settings in Table~\ref{tab:ablation_gamma}. The results indicate that a larger $\gamma$ may lead to better performance on POPE. However, the performance on CHAIR is more sensitive to the increase of $\gamma$, resulting in a decline in performance.
Setting $\gamma$ to $2$ as adopted in our method achieves a better tradeoff between different tasks.

\begin{table}[ht]
\centering
\small
\begin{tabular}{ccccc}
\toprule
\multirow{2}{*}{$\gamma$}        &   \multicolumn{2}{c}{\textbf{POPE}}                                   & \multicolumn{2}{c}{\textbf{CHAIR}}            \\
                                         & \multicolumn{1}{c}{Accuracy$\uparrow$} & \multicolumn{1}{c}{F1 Score$\uparrow$} & \multicolumn{1}{c}{$\text{CHAIR}_i$$\downarrow$} & \multicolumn{1}{c}{$\text{CHAIR}_s$$\downarrow$ }                   \\ \midrule
1.5  & 85.73  &  85.66 &  47.5   &  12.9  \\ 
2    & 86.04  & 85.70 & 47.0  &  12.7  \\ 
3  &   86.32 & 85.88 &  48.2  &   13.9 \\    
5 &   86.25 & 85.91 &  49.1  &   14.3 \\  
                               \bottomrule                             
\end{tabular}
\caption{The performance of our method under the different values of $\gamma$ on LLaVA 1.5.}
\label{tab:ablation_gamma}
\vspace{-0.5cm}
\end{table}

\section{Long Sequence Response Examples}
To further validate the effectiveness of our method, we present some cases of long sequence responses in image captioning tasks in Figure~\ref{fig:qualitative_1} and Figure~\ref{fig:qualitative_3}.  These examples illustrate that the proposed CMAC approach effectively mitigates hallucinations in the generated responses, providing more accurate and contextually relevant descriptions compared to baseline methods.

\section{Limitations}
Our method notably enhances the inference performance of LVLMs by effectively addressing hallucination issues. In the IMD module, we mitigate spurious inter-modality correlations by selectively masking value vectors based on the magnitude of attention weights. However, the magnitude of attention weights may not fully capture the causal relevance between text and visual tokens. Developing a more refined selection mechanism could enable a more accurate estimation of distorted distributions, further improving performance. Additionally, since CMAC does not enhance the visual encoder’s ability to extract and represent relevant knowledge from images, its effectiveness remains constrained by the grounding capability of the visual encoder. Future work will explore these directions to achieve more robust and comprehensive improvements.

% WARNING: do not forget to delete the supplementary pages from your submission 
% \input{sec/X_suppl}

\end{document}